%% file: arxiv.tex
\documentclass[lettersize,journal]{IEEEtran}
\usepackage{cite}
\usepackage{amsmath,amssymb,amsfonts}
\usepackage{algorithmic}
\usepackage{graphicx,color}
\usepackage{textcomp}
\usepackage{xcolor}
\usepackage{mathtools}
\usepackage{algorithm,algorithmic}
\usepackage{makecell} 
\usepackage{booktabs}

\def\BibTeX{{\rm B\kern-.05em{\sc i\kern-.025em b}\kern-.08em
    T\kern-.1667em\lower.7ex\hbox{E}\kern-.125emX}}
\usepackage{balance}

\usepackage{url}

\usepackage{tikz}
    \usetikzlibrary{shapes.arrows}

\usetikzlibrary{positioning}
\usepackage{pgfplots}
\pgfplotsset{
compat=1.3,
legend style={font=\footnotesize, fill opacity=0.7,  draw opacity=1, text opacity=1, draw=white!15!black, legend cell align=left, align=left},
width=6cm,
height=6cm,
yminorticks=false,
xminorticks=false,
title style={font=\small},
tick style={color=black},
tick label style={font=\small},
grid style={line width=.1pt, draw=gray!20},
major grid style={line width=.1pt,draw=gray!20},
}
\usepgfplotslibrary{colorbrewer}
\usepgfplotslibrary{groupplots}
\usepgfplotslibrary{fillbetween}
\usetikzlibrary{plotmarks}
\usetikzlibrary{patterns}
\pgfplotsset{every tick label/.append style={font=\footnotesize}}

\usepackage{subfig}

\usepackage{glossaries}
\input{acronyms.tex}
\usepackage{hyperref}
\newcommand{\E}[1]{\mathbb{E}\left[ #1 \right]} 
\newcommand{\mc}[1]{\mathcal{#1}}   
\DeclareMathOperator*{\argmin}{arg\,min}    

\def \sqwidth{0.23\linewidth}
\def \sqwidthtwo{0.2\linewidth}
\def \sqwidththree{0.23\linewidth}
\def \tabheight{0.15\linewidth}

\def \cwidth{0.95\columnwidth}
\def \cheight {0.45\columnwidth}

\definecolor{color4}{HTML}{FFD700}
\definecolor{color3}{HTML}{EA5F94}
\definecolor{color2}{HTML}{CD34B5}
\definecolor{color1}{HTML}{9D02D7}
\definecolor{color0}{HTML}{0000FF}
\definecolor{darkblue}{HTML}{00429D}
\definecolor{darkgreen}{HTML}{005c00}
\definecolor{gold}{HTML}{D4AF37}
\definecolor{darkred}{HTML}{910000}
\definecolor{darkslategray38}{RGB}{38,38,38}

\definecolor{vir6}{RGB}{253, 231, 37}
\definecolor{vir5}{RGB}{144,215,67}
\definecolor{vir4}{RGB}{53,183,121}
\definecolor{vir3}{RGB}{33,145,140}
\definecolor{vir2}{RGB}{49,104,142}
\definecolor{vir1}{RGB}{68,57,131}
\definecolor{vir0}{RGB}{68, 1, 84}

\definecolor{c0}{RGB}{13,8,135}
\definecolor{c1}{RGB}{106, 0, 168}
\definecolor{c2}{RGB}{177, 42, 144}
\definecolor{c3}{RGB}{225, 100, 98}
\definecolor{c4}{RGB}{252, 166, 54}
\definecolor{c5}{RGB}{240, 249, 33}

\newcommand{\edit}[1]{\textcolor{black}{#1}}

\makeatletter
\newcommand*\titleheader[1]{\gdef\@titleheader{#1}}
\AtBeginDocument{%
  \let\st@red@title\@title
  \def\@title{%
    \bgroup\normalfont\large\centering\@titleheader\par\egroup
    \vskip1.5em\st@red@title}
}
\makeatother

\titleheader{ \vspace{-1.25cm} \small This paper has been submitted to IEEE Transactions on Communications. Copyright may change without notice. \vspace{-.75cm}}

\title{Effective Communication with Dynamic Feature Compression}

\begin{document}


\author{Pietro Talli,~\IEEEmembership{Student~Member,~IEEE}, Francesco Pase,~\IEEEmembership{Graduate~Student~Member,~IEEE},\\ Federico Chiariotti,~\IEEEmembership{Member,~IEEE}, Andrea Zanella,~\IEEEmembership{Senior~Member,~IEEE}, and Michele Zorzi,~\IEEEmembership{Fellow,~IEEE}
\thanks{Pietro Talli (corresponding author, pietro.talli@phd.unipd.it), Francesco Pase (pasefrance@dei.unipd.it), Federico Chiariotti (chiariot@dei.unipd.it), Andrea Zanella (zanella@dei.unipd.it) and Michele Zorzi (zorzi@dei.unipd.it) are with the Department of Information Engineering, University of Padova, 35131 Padua, Italy.}
\thanks{This work was supported by the European Union under the Italian National Recovery and Resilience Plan of NextGenerationEU, under the partnership on ``Telecommunications of the Future'' (PE0000001 - program
``RESTART'') and the ``SoE Young Researchers'' grant REDIAL.}}

\maketitle

\begin{abstract}
The remote wireless control of industrial systems is one of the major use cases for 5G and beyond systems: in these cases, the massive amounts of sensory information that need to be shared over the wireless medium may overload even high-capacity connections. Consequently, solving the \emph{effective communication} problem by optimizing the transmission strategy to discard irrelevant information can provide a significant advantage, but is often a very complex task. In this work, we consider a prototypal system in which an observer must communicate its sensory data to a robot controlling a task (e.g., a mobile robot in a factory). We then model it as a remote \gls{pomdp}, considering the effect of adopting semantic and effective communication-oriented solutions on the overall system performance. We split the communication problem by considering an ensemble \gls{vqvae} encoding, and train a \gls{drl} agent to dynamically adapt the quantization level, considering both the current state of the environment and the memory of past messages. We tested the proposed approach on the well-known CartPole reference control problem, obtaining a significant performance increase over traditional approaches.
\end{abstract}

\begin{IEEEkeywords}
Effective communication, Networked control, Semantic communication, Information bottleneck
\end{IEEEkeywords}


\section{Introduction}
\label{sec:intro}

\IEEEPARstart{T}{he} main goal of classical communication theory is to build reliable systems for the accurate transmission of \edit{arbitrary} data \edit{through a constrained communication channel while using as few symbols as possible}. However, in the preface to Shannon's seminal work~\cite{shannon1949mathematical}, Warren Weaver already envisioned two more complex Levels of communication beyond the simple transmission of bits. Classical communications are then included in Level A, or the \emph{technical problem}, which concerns itself with the accurate and efficient transmission of arbitrary raw data. Level B, or the \emph{semantic problem}, is to find the best way to convey the meaning of the message, even when irrelevant details are lost or misunderstood, while Level C, also called the \emph{effectiveness problem}, deals with the resulting behavior of the receiver~\cite{gunduz2022beyond}: as long as the receiver takes the optimal decision, the effectiveness problem is solved, regardless of the quality of the received information. \edit{The Level B and C problems are tightly intertwined, as defining the meaning of a message is often related to the intentions of the receiver.}

While the Level B and C problems attracted limited attention for decades, the explosion of \gls{iiot} systems has drawn the research and industrial communities toward semantic and effective communication~\cite{popovski2020semantic}, optimizing remote control processes under severe communication constraints beyond Shannon's limits on Level A performance~\cite{gunduz2022beyond}. In particular, the effectiveness problem is highly relevant to robotic applications, in which independent mobile robots, such as drones or rovers, must operate based on information from remote sensors. In this case the sensors and the cameras act as the transmitter in a communication problem, while the robot is the receiver: by solving the Level C problem, the sensors can transmit the information that best directs the robot's actions toward the optimal policy~\cite{stavrou2022rate}. We can also consider a case in which the robot is the transmitter, while the receiver is a remote controller, which must get the most relevant information to decide the control policy~\cite{wan2020cognitive}.

The rise of communication metrics that take the content of the message into account, such as the \gls{voi}~\cite{yates2020agesurvey}, represents an attempt to approach the problem in practical scenarios, and analytical studies have exploited information theory to define a semantic accuracy metric and minimize distortion~\cite{shao2022theory}. In particular, \emph{information bottleneck} theory~\cite{beck2022semantic} has been widely used to characterize Level B optimization~\cite{shao2021learning}. However, translating a practical system model into a semantic space is a non-trivial issue, and the semantic problem is a subject of active research~\cite{uysal2022semantic, pase2023semantic}.
The effectiveness problem is even more complex, as it implicitly depends on estimating the effect of communication distortion on the control policy and, consequently, on its performance~\cite{tung2021effective}. While the effect of simple scheduling policies is relatively easy to compute~\cite{kim2019learning}, and linear control systems can be optimized explicitly~\cite{zheng2020urgency}, realistic control tasks are highly complex, complicating an analytical approach to the Level C problem. Pure learning-based solutions that consider communication as an action in a multi-agent \gls{drl} problem, such as emergent communication, also have limitations~\cite{foerster2016learning}, as they can only deal with very simple scenarios due to significant convergence and training issues. In some cases, the information bottleneck approach can also be exploited to determine state importance~\cite{goyal2018transfer}, but the existing literature on optimizing Level C communication is very sparse, and limited to simpler scenarios~\cite{rcmab_pase}. \edit{Another possible approach to the remote tracking of Markov sources is addressed in zero-delay coding theory \cite{akyol2014zero}. However, this theory considers the error on the hidden state estimate as the objective of the optimization, which is similar to what we could consider a semantic (or Level B) approach. In this work, we show that it is possible to optimize the system with respect to other metrics which cannot be explicitly derived such as cumulative rewards in \gls{drl}: by accepting a higher distortion at the semantic Level when it is not relevant to the task, we can further reduce the required bitrate without sacrificing the control performance.}

In this work, we consider a dual model which combines concepts from \gls{drl} and semantic source coding: we consider an ensemble of \gls{vqvae} models~\cite{van2017neural}, each of which learns to represent observations using a different codebook. A \gls{drl} agent can then select the codebook to be used for each transmission, controlling the trade-off between accuracy and compression. Depending on the task of the receiver, the reward to the \gls{drl} agent can be tuned to solve the Level A, B, and C problems, optimizing the performance for each specific task. In order to test the performance of the proposed framework in a relevant example scenario, we consider the well-known CartPole problem, whose state can be easily converted into a semantic one, as its dynamics depend on a limited set of physical quantities. \edit{The problem we selected is purposefully simple, as this allows for a better explainability and an easier training, but the solution is not limited to the CartPole problem, and can be adapted to more complex tasks.} The main contributions of this paper are then given by the following:
\begin{itemize}
\item We model a remote-control system as a remote \gls{pomdp} problem and present an efficient solution for learning effective communication through the dynamic compression of learnable features;
\item We show that dynamic codebook selection outperforms static strategies for all three Levels, and that considering the Level C task can significantly improve the control performance without increasing the bitrate;
\item We adopt an explainability framework to understand the choices of the agent in this simple problem, and verify that the Level C dynamic compression captures the receiver's uncertainty in the state estimation and its impact on the expected reward, transmitting only when necessary.
\end{itemize}
\edit{We remark that the dynamic codebook selection policy is not limited to the \gls{vqvae} ensemble we consider, but is a general technique that can be applied to any compression algorithm with adaptable quality parameters, helping deliver more accurate information when it is relevant to do so.}
The results and policy analysis lead to significant insights for the design of communication strategies for remote control. A partial version of this work was presented at the IEEE INFOCOM WiSARN 2023 workshop~\cite{talli2023infocom}. This paper includes a more complete theoretical characterization of the problem, as well as an updated learning architecture, additional results, and an in-depth analysis of the \gls{drl}-based dynamic compression policy.

The rest of the paper is organized as follows: first, we analyze the state of the art on semantic and effective communication in Sec.~\ref{sec:related}. Sec.~\ref{sec:system} then presents the general system model and the three Levels of communication we consider. We then describe the dynamic feature compression solution in Sec.~\ref{sec:solution}, which is evaluated by simulation in Sec.~\ref{sec:results}. Finally, Sec.~\ref{sec:conclusion} concludes the paper.

\section{Related Work}
\label{sec:related}

The specific requirements of distributed and remotely controlled systems have focused the research community's attention towards communication systems that must provide updated information to enable real-time high-level tasks such as inference, tracking or control. Although metrics such as \gls{aoi} \cite{yates2020agesurvey} represent a major improvement with respect to latency and packet loss, they are still limited, as they assume that the quality of the information available at the receiver degrades deterministically with time, most commonly (but not necessarily~\cite{kosta2021age}) in a linear fashion. However, more sophisticated systems can also take into consideration the current state of the system in order to decide whether and when to update the status of the receiver. Metrics such as \gls{uoi}~\cite{zheng2020urgency} and \gls{voi}~\cite{wang2022framework} incorporate state information in their definition and are thus aware of the intrinsic value of potential updates. Other context-aware indices to measure the nonlinear time-varying importance and the non-uniform context dependence of the status information have also been proposed~\cite{zheng2020urgency}. The authors of~\cite{fountoulakis2023goal} considered a system in which a transmitter monitors the status of a system and updates the controller, providing status information. Then a constrained \gls{mdp} is formulated to minimize the cost of actuation and simultaneously guarantee a target communication rate. Both works show significant improvements with respect to other metrics such as \gls{aoi}.

At the same time, the development of learning-based coding schemes has allowed communication system designers to move beyond packet error as the key coding performance metric, exploiting semantic considerations. Joint source-channel coding for wireless image transmission is implemented in~\cite{deepjscoding,Huiqiang_semantic}, and the encoder-decoder pair is parameterized by a \gls{nn}, whose architecture may vary. This approach can be used to maintain the semantic information contained in the transmitted data, while improving the compression performance.

Semantic information at the receiver can be used to solve different tasks. Effective communication \cite{tung2021effective} can be seen as an extension of this, in which the task involves the receiver taking actions and possibly altering the information that the transmitter is communicating. 
Effective communication differs from semantic communication mostly because the ``semantic'' content which has to be preserved in the communicated messages is not explicit. Moreover, control tasks have a temporal component that must be taken into account, as investigated in \cite{tung2021effective}. The scenario considered in the work is a two-agent \gls{pomdp} in which one agent communicates and the other agent interacts with the environment, using \gls{drl} to solve the joint problem and encoding the information. A distributed perception scenario, in which multiple sensors communicate to a single robot, is considered in~\cite{mason2023multi} and solved using \gls{marl}, showing that joint training improves the performance of the system, particularly when communication is severely constrained. While past works aimed at specific scenarios and objectives, this paper proposes a novel \gls{drl} approach that combines status updates with an adaptive coding scheme and can be easily adapted to operate on any of the three Levels of communication (A, B, or C).

\section{System Model}
\label{sec:system}

The recent interest in semantic and effective communications from the research community has driven the development of a wide array of models and conceptualizations, as highlighted in the previous section. At the highest level of abstraction, our purpose is to define a model in which effective communication is meaningful, and the differences between the three problems in Weaver's formulation become clear.

Let us then consider a simple example: we have a remote actuator performing a control task, while a camera observes the results and transmits its observation through a wireless channel. The actuator might have its own sensors, but it relies on the video feed to improve its performance and maintain stable and efficient control. The classical, Level A approach to the problem would be to compress the video as efficiently as possible, minimizing the reconstruction error on frames by using an appropriate codec. 
The difference between Level A and Level B solutions is then obvious: the former encodes new frames so that the reconstruction fidelity is preserved, while the latter maps elements in the frame to their importance when estimating the physical state of the system. In some control applications, the state can be defined trivially, while in others it may be more complex, but in general, the translation of the video to the state space is unaffected by irrelevant information (such as, e.g., movements in the background).

If we consider Level C, we target control performance directly, and thus further restrict the definition of relevant information: while Level B concerns itself with estimating the system state correctly, a Level C solution only considers errors in the state estimation when they cause performance drops. If the control action is the same over a wide set of states, accuracy then becomes unnecessary, as the actuator only needs a rough estimate of the state to decide what to do; the same happens if there are multiple actions with almost equivalent performance, i.e., if the optimality gap caused by imperfect information remains small.

These natural observations represent the core concepts of effective communication, but implementing them in practical systems is often complex as actions have long-term consequences, and state estimates are based on a history of observations, so that transmitting a message may affect future performance in complex ways. In the following, we provide an analytical framework using the \emph{remote \gls{pomdp}} approach to objectively evaluate these choices and implement a solution for effective communication in cyber-physical systems. We will denote random variables with capital letters, their possible values with lower-case letters, and sets with calligraphic or Greek capitals. Table~\ref{tab:symbols} reports the main symbols we introduce in the following sections for the reader's convenience. 

\begin{table}[]
    \centering
    \caption{Main notation and definitions.}
    \begin{tabular}{cl}
        \toprule
        Symbol & Definition \\
        \midrule
        $\mathcal{S}$ & Set of system states \\
        $\mathcal{A}$ & Set of feasible actions \\
        $\mathcal{O}$ & Set of observations \\
        $P$ & State transition probability function \\
        $\omega$ & Observation function \\
        $R$ & Reward function \\
        $\gamma$ & Discount factor \\
        $h$ & History of stochastic observations \\
        $h^{(r)}$ & History of received messages at the robot \\
        $\pi$ & Policy \\ 
        $G$ & Expected cumulative discounted reward \\
        $\Phi(\cdot)$ & Space of probability distributions over a set \\
        $\xi$ & Belief distribution over the state space $\mathcal{S}$ \\
        $\xi^{(o)}$ & Belief distribution at the observer \\
        $\xi^{\text{pri,}(r)}$ & Prior belief distribution at the robot \\
        $\xi^{(r)}$ & Belief distribution at the robot \\
        $\mc{M}$ & Set of messages\\
        $\Lambda$ & Encoding function \\ 
        $m$ & Message communicated \\
        $\ell(m)$ & Length of message $m$ \\
        $\zeta$ & Vector quantizer \\
        $\mc{P}$ & Picture space \\ 
        $\beta$ & Communication cost \\
        
        \bottomrule
    \end{tabular}
    
    \label{tab:symbols}
\end{table}

\subsection{POMDP Definition and Solution}


In the \edit{infinite horizon} \gls{pomdp} formulation~\cite{pomdp}, one agent needs to optimally control a stochastic process defined by a tuple $\left<\mc{S}, \mc{A}, \mc{O}, P, \omega, R, \gamma\right>$, where $\mc{S}$ represents the set of system states, $\mathcal{A}$ is the set of feasible actions, and $\mc{O}$ is the observation set. The function $P : \mc{S} \times \mathcal{A} \rightarrow \Phi(\mc{S})$, where $\Phi(\cdot)$ represents the space of probability distributions over a set, gives the state transition probability function. We denote the conditional probability distribution of the next state, given the current state and the selected action, as $P(s'|s,a)=\Pr\left[S_{t+1}=s'|S_t=s,A_t=a\right]$. Then, $\omega : \mc{S} \rightarrow \Phi(\mc{O})$ is an observation function \edit{representing the probability of an observation conditioned on the system state:} $\omega(o|s) = \mathrm{Pr} \left[ O_t = o| S_t = s\right]$, i.e., the probability of receiving observation $o$, given that the system is in state $s$. Finally, function $R: \mc{S} \times \mathcal{A} \rightarrow \mathbb{R}$ \edit{represents} the expected reward received by the agent when taking action $a$ in state $s$, denoted as $R(s,a)$, and the scalar $\gamma \in [0, 1)$ is a discount factor used to compute the long-term reward. We notice that functions $P$, $R$, and $\omega$ do not depend on the time instant $t$, thus focusing on homogeneous Markov processes.

The \gls{pomdp} proceeds in discrete steps indexed by $t$: at each step $t$, the agent can infer the system state $s_t$ only from the partial information given by the history of stochastic observations $h_t=(o_t, o_{t-1},\ldots,o_1) \in \mc{O}^t$. Based on these observations, and on its policy $\pi : \mc{O}^t \rightarrow \Phi(\mathcal{A})$, which outputs a probability distribution over the action space for each possible observation history, the agent interacts with the system by selecting an action ${A_t \sim \pi(h_t)}$. The sampled action $a_t$ is then performed in the real environment, whose hidden state $s_t$ is unknown to the agent, which then receives a feedback from the environment in the form of a (potentially stochastic) reward $r_t$, with expected value $R_t=R(s_t,a_t)$.\footnote{As the reward signal is only provided during training, it cannot be used to infer the value of $s_t$.} The goal for the agent is then to optimize its policy $\pi$ to maximize the expected cumulative discounted reward $G = \mathbb{E}\Big[ \sum_{t} \gamma^t R_t \Big]$. 
Having an optimal policy is equivalent to knowing the optimal state-action values, also known as $Q$-values, and taking action 
\begin{equation*}
a_t = \pi(h_t) = \underset{a \in \mathcal{A}}{\arg\max} \ Q(h_t, a),
\end{equation*}
where \edit{$Q(h_t, a) = \E{\sum_{\tau=t}^{\infty}\gamma^{\tau-t}r_{\tau}|h_t,a}$ is the expected cumulative reward starting from $(h_t,a)$}. 

However, considering the full history of observations makes the solution highly complex, as the length of $h_t$ is potentially unbounded. We then define an estimator $\xi:\mc{O}^t\rightarrow\Phi(\mc{S})$, which outputs the \emph{a posteriori} belief distribution over the state space. We can then recast the original \gls{pomdp} as a standard \gls{mdp}, whose state space is $\mc{S}'=\Phi(\mc{S})$, i.e., the space of possible belief distributions. Solving the \gls{pomdp} in this modified belief space has been proved to be optimal in~\cite{sondik1971optimal}.

The policy over this modified \gls{mdp} is then $\pi:\mc{S}'\rightarrow\mc{A}$, which can be optimized using standard tools~\cite{sutton2018reinforcement}. We can also compute the new transition probability and expected reward as in~\cite{sondik1971optimal}.
Given $\xi_t(s) = \Pr\left(S_t=s\mid h_t\right)$\edit{, which represents the maximum likelihood estimate of the state given all the history of observations $h_t$,} and $A_t=a$, 
we define the \emph{a priori} belief over the state at time $t+1$ as 
\begin{equation*}
\begin{aligned}
    \xi^{\text{pri}}_{t+1}\left(s|\xi_t,a\right)=&\Pr\left[S_{t+1} = s \mid \xi_t, A_t=a\right] \\
    =& \sum_{s' \in \mc{S}} \xi_t(s') P(s|s',a).
\end{aligned}
\end{equation*}
The \emph{a posteriori} belief $\xi_{t+1}$\edit{, which also includes the new observation $o_{t+1}$,} can then be obtained by performing a Bayesian update using the \emph{a priori} belief as a prior:
\begin{equation}\label{eq:bayesian}
\begin{aligned}
    \xi_{t+1}\left(s|\xi_{t},a,o\right)=&\Pr\left[S_{t+1} = s \mid \xi_t, A_t=a,O_{t+1}=o\right]\\
    =&\Pr\left[S_{t+1} = s \mid \xi^{\text{pri}}_{t+1}\left(s|\xi_t,a\right),O_{t+1}=o\right]\\
    =& \frac{\omega(o|s)\xi^{\text{pri}}_{t+1}(s|\xi_t,a)}{\sum_{s'\in\mc{S}}\xi^{\text{pri}}_{t+1}(s'|\xi_t,a)\omega(o|s')}.
\end{aligned}
\end{equation}
These update equations allow us to compute the modified transition probability matrix $P'$ for the belief \gls{mdp}, which is then defined by the tuple $\left<\Phi(\mc{S}), \mc{A}, P', R, \gamma\right>$.

\subsection{The Remote POMDP}

In this paper we consider a variant of the \gls{pomdp}, that we define \emph{remote \gls{pomdp}}, in which two agents are involved in the process. The first agent,  i.e., the \emph{observer}, receives observation $O_t \in \mc{O}$, and needs to convey such information to a second agent, i.e., the \emph{robot}, through a constrained communication channel, which limits the number of bits the observer can send reliably. Consequently, the amount of information the observer can send to the robot is limited. The robot then chooses and takes an action in the physical environment. This system can formalize many control problems in future \gls{iiot} systems, as sensors and actuators may potentially be geographically distributed, and the amount of information they can exchange to accomplish a task is limited by the shared wireless medium, which has to be allocated to the many devices installed in the factory, as well as by the energy limitations of the sensors. Similar systems have been analyzed in \cite{tung2021effective, fountoulakis2023goal}. We will now analyze the problems for the two agents, considering a case in which communication and control are designed separately. Joint control and communication approaches~\cite{mason2023multi} can outperform separate approaches by tuning the two agents' policies to each other, but they introduce additional training complexity, and will not be considered in this work. In the following, we will refer to variable $x$ related to the robot as $x^{(r)}$, while the corresponding variable on the observer side will be denoted by $x^{(o)}$.

\subsubsection{The Robot-Side POMDP}
We denote the message communicated to the robot at time step $t$ as $m_t \in \mc{M}$. The set of possible messages $\mc{M}$ forms the set of observations that are available to the robot, and the history of these observations is given by $h^{(r)}_t = \lbrace m_t,\ldots,m_1\rbrace$, which is the sequence of messages received up to time $t$. We can then see the robot as an agent with its own \gls{pomdp}, in which the observations are filtered by both the partial knowledge of the observer and the further distortion produced by the fact that these observations are encoded and communicated through a constrained channel. The robot-side \gls{pomdp} is then defined by the tuple $\left<\mc{S}, \mc{A}, \mc{M}, P, \pi^{(o)}, R, \gamma\right>$, as observations depend on the observer's policy. 

The message transmitted from the observer to the robot modifies the belief distribution over the next state as a Bayesian update. Let us define the distribution over the current state, given that the message $m_{t}$ has been received, as $\xi^{(r)}_t$. For example, if the communicated message $m_{t}$ contains the correct state $S_t=s$, the belief distribution becomes deterministic, i.e., $\xi(s' \mid m_{t}) = \delta_{s,s'}$, where $\delta_{m,n}$ is the Kronecker delta function, equal to 1 if the two arguments are the same and 0 otherwise. Ideally, an intelligent observer will allocate more communication resources and thus provide more precise messages if the \emph{a priori} distribution of the robot is far from the one estimated by the observer. The modified \gls{mdp} is then defined by the tuple 
$\left<\Phi(\mc{S}), \mc{A}, P^{(r)}, R, \gamma\right>$, where $P^{(r)}$ represents the Bayesian update function. According to the previous notation, we can express the optimal action at time $t+1$ as 
\begin{equation*}
    a_{t+1} = \underset{a \in \mathcal{A}}{\arg \max} \ Q\left(\xi^{(r)}_t,a\right),
\end{equation*}
where $\xi^{(r)}_t$ is the current belief at the robot side (after message $m_t$ is received). The robot's reward is simply given as the reward of the original \gls{pomdp}, i.e., the control performance in the environment. The optimal policy can be reached by using standard \gls{drl} tools. 

\subsubsection{The Observer-Side POMDP} On the other side, the observer needs to encode its belief $\xi_t^{(o)}$ in a message $m_t\in\mc{M}$ and transmit it. We can then consider the observer-side \gls{pomdp}, in which the action set corresponds to the set of messages $\mc{M}$ and the state space is represented by the belief from the observed results. The tuple defining this \gls{pomdp} is $\left<\mc{S}, \mc{M}, \mc{O}, P, \omega, R^{(o)}, \gamma\right>$. We assume that the observer knows the robot's policy, i.e., it can know the actions that the robot takes in the environment and use them to improve its estimate of the state. This can also be accomplished if the robot transmits the actions it takes as feedback to the observer. As described above for the robot-side problem, we can transform this \gls{pomdp} into the belief \gls{mdp} given by $\left<\Phi(\mc{S})\times\Phi(\mc{S}), \mc{M}, P^{(o)}, R^{(o)}, \gamma\right>$, where $P^{(o)}$ is the Bayesian update given in~\eqref{eq:bayesian}. We highlight that the observer needs to keep track of both its own and the robot's belief, as the effectiveness of communication depends on the difference between the two, and the state of the observer is given by $\left<\xi^{(o)}_t,\xi^{\text{pri},(r)}_t\right>$.

The objective of the observer is to minimize channel usage, i.e., communicate as few bits as possible, while maintaining the highest possible performance in the control task: the expected reward $R^{(o)}$ depends on both components. \edit{We then consider a simple linear combination approach:} if it transmits message $m$, whose length in bits is $\ell(m)$, the observer then gets a penalty $\beta\ell(m)$, where $\beta\in\mathbb{R}^+$ is a cost parameter. In order to optimize its policy, the observer also needs to have a way to gauge the \emph{value} of information, which is a complex problem: information theory, and in particular rate-distortion theory, have provided the fundamental limits when optimizing for the technical problem, i.e., Level A, where the goal is to reconstruct the source signals with the highest fidelity~\cite{cover:IT}. We will discuss the definition of \gls{voi} in the following sections. \edit{More complex modeling choices for the transmission cost are also possible, e.g., considering energy constraints for a sensor with energy harvesting capabilities, but are beyond the scope of this work.}

As the complexity of the problem is massive, we restrict ourselves to a smaller action space by making a simplifying assumption, which allows us to separate the problem: the observer does not transmit the entire belief distribution, which may be implicit, but rather the observation $O_t$. We then consider the encoding function $\Lambda: \mc{O}\times\Phi(\mc{S}) \rightarrow \mc{M}$, which will generate a message $M_t = \Lambda\left(O_t\mid \xi^{(o)}_t\right)$ to be sent to the robot at each step $t$.

\subsection{Observer Reward in Remote POMDPs}\label{ssec:coding}

The first and simplest way to solve the remote \gls{pomdp} problem is to blindly apply standard Level A rate-distortion metrics to compress the sensor observations into messages to be sent to the agent. As an example, in the CartPole problem analyzed in this work (see Sec.~\ref{sec:results}), one sensor observation is given by two consecutive 2D camera acquisitions. The observer's policy is then independent of the robot's task, and can be computed separately. The Level A reward function $R_A^{(o)}$ is then given as follows:
\begin{equation}
R_A^{(o)}\left(\left<\xi^{(o)}_t,\xi^{\text{pri},(r)}_t\right>,m\right)=-d_A(o_t,\hat{o}_t)-\beta\ell(m).\label{eq:rew_A}
\end{equation}
In the CartPole case, a natural distortion metric is the image \gls{psnr}, an image quality metric proportional to the logarithm of the normalized \gls{mse} between the images. Naturally, encoding the observation with a higher precision will require more bits, as the set of messages needs to be bigger.

The Level B problem considers the projection of the raw observations into a significantly smaller \emph{semantic space}, over which we measure distortion using function $d_B$, explicitly capturing the error over the needed physical system information, e.g., the angular position and velocity of the pole in the CartPole problem. The Level B reward function $R_B^{(o)}$ is then given as follows:
\begin{equation}
R_B^{(o)}\left(\left<\xi^{(o)}_t,\xi^{\text{pri},(r)}_t\right>,m\right)=-d_B\left(\xi^{(o)}_t,\xi^{(r)}_t\right)-\beta\ell(m).\label{eq:rew_B}
\end{equation}
In our CartPole case, this may be simply represented by the \gls{mse} between the best estimate of the state at the transmitter and the receiver.

Finally, we can consider the Level C system. In this case, the distortion metric is not needed, as the control performance can be used directly, and the reward $R_C^{(o)}$ is:
\begin{equation}
R_C^{(o)}\left(\left<\xi^{(o)}_t,\xi^{\text{pri},(r)}_t\right>,m\right)=R\left(\xi^{(r)}_t,\pi^{(r)}\left(\xi^{(r)}_t\right)\right)-\beta\ell(m).\label{eq:rew_C}
\end{equation}
The \gls{voi} of message $m$, $V\left(\xi^{\text{pri},(r)}_t,m\right)$, can then be given by the difference between the expected performance of the robot with and without this information:
\begin{equation}\label{eq:voi_C}
\begin{aligned}
V\left(\xi^{\text{pri},(r)}_t,m\right)=&Q\left(\xi^{(r)}_t,\pi^{(r)}\left(\xi^{(r)}_t\right)\right)\\
&-Q\left(\xi^{\text{pri},(r)}_t,\pi^{(r)}\left(\xi^{\text{pri},(r)}_t\right)\right).
\end{aligned}
\end{equation}
Thus, the optimal Level C observer policy $\pi_C^{(o)}$ will balance the trade-off between the performance at the receiver and the communication cost not only in the current time step but also in the long term. This foresighted behavior is essential when considering that the belief distributions incorporate the memory of previously received messages. Providing information that does not improve the expected reward in the next step might still be worth the cost if it allows the robot to improve its estimate, reducing the need for future communication.

\section{Proposed Solution}
\label{sec:solution}
 In this section, we introduce the architecture we used to represent $\Lambda$, the \gls{vqvae}, and discuss the remote \gls{pomdp} solution. As the \gls{vqvae} model is not adaptive, we consider an ensemble model with different quantization levels, limiting the choice of the observer to \emph{which} \gls{vqvae} model to use in the transmission. As we mentioned, directly learning the encoding is highly complex, with a vast action space, and techniques such as emergent communication that learn it explicitly are limited to scenarios with very simple tasks and immediate rewards. By restricting the problem to a smaller action space, we \edit{find a potentially sub-optimal solution, but we can deal with much more complex problems}.

\subsection{Deep VQ-VAE Encoding}
\label{subsec:vae}
In order to represent the encoding function $\Lambda$, and to restrict the observer-side \gls{pomdp} to a more manageable action space, the observer exploits the \gls{vqvae} architecture introduced in~\cite{van2017neural}. The \gls{vqvae} is built on top of the more common \gls{vae} model, with the additional feature of finding an optimal discrete representation of the latent space. The \gls{vae} is used to reduce the dimensionality of an input vector $X \in \mathbb{R}^I$, by mapping it into a stochastic latent representation $Z \in \mathbb{R}^L \sim q_{\nu}(Z | X)$, where $L < I$. The stochastic encoding function $q_{\nu}(Z | X)$ is a parameterized probability distribution represented by a neural network with parameter vector $\nu$. To find optimal latent representations $Z$, the \gls{vae} jointly optimizes a decoding function $p_{\theta}(\hat{X}|Z)$ that aims to reconstruct $X$ from a sample $\hat{X} \sim p_{\theta}(\hat{X} | Z)$. This way, the parameter vectors $\nu$ and $\theta$ are usually jointly optimized to minimize the distortion $d(X, \hat{X})$ between the input and its reconstruction, given the constraint on $Z$, while reducing the distance between $q_{\nu}(Z|X)$, and some prior $q(Z)$~\cite{vae2013} used to impose some structure or complexity budget.

However, in practical scenarios, one needs to digitally encode the input $X$ into a discrete latent representation. To do this, the \gls{vqvae} quantizes the latent space by using $N$ $K$-dimensional codewords $z_1, \dots, z_N \in \mathbb{R}^K$, forming a dictionary with $N$ entries. Moreover, to better represent 3D inputs, the \gls{vqvae} quantizes the latent representation $Z$ using a set of $F$ blocks, each quantizing one feature $f(X)$ of the input, and chosen from a set of $N$ possible codewords. We denote the set containing all the $N^F$ possible concatenated blocks with $\mathcal{M}(N)$, as it represents the set of all possible messages the observer can use to convey to the robot the information on the observation $O$, by using $F$ discrete $N$-dimensional features. The peculiarity of the \gls{vqvae} architecture is that it jointly optimizes the codewords in $\mathcal{M}(N)$ together with the stochastic encoding and decoding functions $q_{\nu}$ and $p_{\theta}$, instead of simply applying fixed vector quantization on top of learned continuous latent variables $Z$. When the communication budget is fixed, i.e., the value of $L$ is constant, the protocol to solve the remote \gls{pomdp} is rather simple: first, the observer trains the \gls{vqvae} with $N = 2^{F^{-1}L}$ to minimize the technical, semantic, or effective distortion $d_{\alpha}$, depending on the problem; then, at each step $t$, the observer computes $\hat{m} \sim q_{\nu}(\cdot | o_t)$, and finds $ m_t = \argmin_{m \in \mathcal{M}(N)} \| m - \hat{m} \|_2$.
The message $m_t$ is sent to the robot, which can optimize its decision accordingly.

\subsection{Dynamic Feature Compression}
\label{subsec:drl}

\begin{figure}
\centering
\input{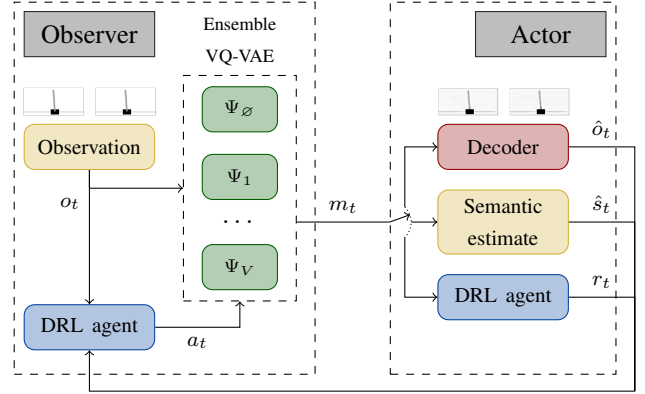}
\caption{Dynamic feature compression architecture.}
\label{fig:arch}
\end{figure}

We can then consider the architecture shown in Fig.~\ref{fig:arch}, consisting of a set of \glspl{vqvae} $\mathcal{V} = \{ \zeta_\varnothing, \zeta_1, \ldots, \zeta_V\}$, where each \gls{vqvae} $\zeta_v$ compresses each feature using $v$ bits. We also include a null action $\zeta_\varnothing$, which corresponds to not transmitting anything. As we only consider the communication side of the problem, the actor is trained beforehand using the messages with the finest-grained quantization, which are compressed with the \gls{vqvae} $\zeta_V$ with the largest codebook. \edit{This choice ensures that the actor can deal with finer-grained inputs, while still being robust to lower-precision features.}
The robot can then perform three different tasks, corresponding to the three communication problems: it can decode the observation (Level A) with the highest possible accuracy, using the decoder part of the \gls{vqvae} architecture; it can estimate the hidden state (Level B) using a supervised learning solution; or it can perform a control action based on the received information and observe its effects (Level C).

In all three cases, the dynamic compression is performed by the observer, based on the feedback from the robot. The observer side of the remote \gls{pomdp}, whose reward is given in~\eqref{eq:rew_A}-\eqref{eq:rew_C}, is restricted to the choice of $\zeta_v$, i.e., to selecting one of the possible codebooks learned by each \gls{vqvae} in the ensemble model, or to avoid any transmission. As we described in the previous section, the type of reward depends on the communication problem that the observer is trying to solve: at Levels A and B, the observer aims at minimizing distortion in the observation and semantic space, respectively. At Level C, the objective is to maximize the robot's reward. \edit{We remark that the only Level at which the decision of the receiver matters is Level C. The semantic estimate describes the physical process, which models the dynamics of the control process that the Observer can sense. Optimizing the transmitter to minimize the reconstruction error in the semantic estimate does not consider the decision of the actor (receiver). In general, the physical state of the system can carry redundant or irrelevant information with respect to the agent's decision, and is not equivalent to Level C optimization.}

In all three cases, \emph{memory} is important: representing snapshots of the physical system in consecutive instants, subsequent observations have high correlations, and the robot can glean a significant amount of information from past messages. This is an important advantage of dynamic compression, as it can adapt messages to the estimated knowledge at the receiver side.

While the observer is adapting its transmissions to the robot's task, the robot's algorithms are fixed. They could themselves be adapted to the dynamic compression strategy, but this joint training is significantly more complex, and we consider it as a possible extension of this work.

\subsection{RL implementation} 
There are two policies in the considered system, one for the observer and one for the robot and in both cases the policies are learned through the Actor Critic algorithm. This means that an agent learns a parametric policy $\pi_{\lambda}$ and a $Q$-values estimator. Both the policy and the $Q$-values are neural networks. In order to take into account the past observations the two networks share a \gls{lstm} layer which estimates a latent state which is then given as input to both the policy and the values estimator. This architecture avoids explicitly modeling the belief distribution which may be complicated to treat in continuous settings like the one considered in this work. This practical choice is also useful to avoid decoding the latent features discovered by the \gls{vqvae} back in the observation space $\mc{O}$ or in the physical state space $\mc{S}$, increasing the potential for errors. Indeed, the quantized features communicated with the message $m_t$ contain a structured representation of the observation space which can be used effectively by an \gls{lstm} to estimate the \emph{true state}. The training algorithm is the standard \gls{a2c}, but the replay buffer is appropriately modified to take into account the history of previously received messages. 

\section{Simulation Settings and Results}
\label{sec:results}

The underlying use case analyzed in this work is the well-known CartPole problem, as implemented in the \emph{OpenAI Gym} library.\footnote{\url{https://www.gymlibrary.dev/environments/classic_control/cart_pole/}} In this problem, a pole is installed on a cart, and the task is to control the cart position and velocity to keep the pole in equilibrium. The physical state of the system is fully described by the cart position $x_t$ and velocity $\dot{x}_t$, and the pole angle $\psi_t$ and angular velocity $\dot{\psi}_t$. Consequently, the true state of the system is ${s_t = ( x_t, \dot{x}_t, \psi_t, \dot{\psi}_t )}$, and the semantic state space is $\mc{S}\subset\mathbb{R}^4$ (because of physical constraints, the range of each value does not actually span the whole real line). \edit{The main simulation parameters are reported in Table~\ref{tab:parameter_table}.}

\begin{figure}[t]
\centering
\subfloat[Original.]{
\includegraphics[width=0.8\columnwidth]{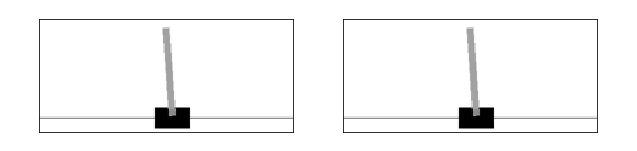}}

\subfloat[Reconstructed.]{\includegraphics[width=0.8\columnwidth]{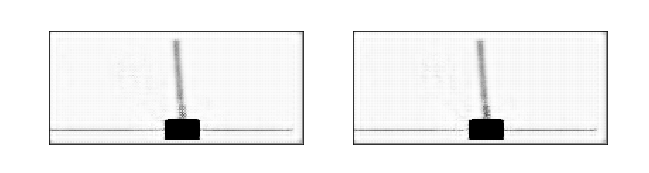}}
\caption{Example of the original and reconstructed observation.}
\label{fig:cartpole}
\end{figure}

At each step $t$, the observer senses the system by taking a black and white picture of the scene, which is in a space $\mc{P}=\left\{ 0, \dots, 255\right\}^{180 \times 360}$. To take the temporal element into account, an observation $O_t$ includes two subsequent pictures, at times $t-1$ and $t$, so that the observation space is $\mc{O}=\mc{P}\times\mc{P}$. An example of the transmission process is given in Fig.~\ref{fig:cartpole}, which shows the original version sensed by the observer (above) and the reconstructed version at the receiver (below) when using a trained \gls{vqvae} $\zeta_{6}$\edit{, i.e., an encoder trained with 6 bits per feature, the maximum we consider in this study}.

\begin{table}[t]
    \centering
    \caption{Simulation Parameters.}
    \begin{tabular}{ccl}
        \toprule
        Parameter & Value & Description \\ 
        \midrule
        $H \times W$ & $160 \times 360$ & Image size \\
        $V$ & 7 & Number of quantizers \\
        $F$ & 8 & Number of latent features \\
        $D_{\text{emb}}$ & 8 & Embedding dimension of features \\
        $B$ & 256 & Batch size \\ 
        $\gamma$ & 0.95 & Discount factor \\
        $T$ & 500 & Maximum number of steps for an episode \\
        $\alpha_{\text{enc}}$ & $10^{-3}$ &  \gls{vqvae} learning rate \\ 
        $\alpha_{\text{Reg}}$ & $10^{-4}$ &  Regressor learning rate \\ 
        $\alpha_{\text{A2C}}$ & $10^{-4}$ & \gls{a2c} learning rate \\ 
        $D$ & $5\times 10^{4}$ & Size of the VQ-VAE training dataset \\ 
        $T_{\text{enc}}$ & $100$ & Encoder training epochs \\ 
        $E_{\text{rob}}$ & $2 \times 10^{4}$ & Robot policy training episodes \\
        $E_{\text{obs}}$ & $1 \times 10^{5}$ & Observer policy training episodes \\ 
        $E_{\text{test}}$ & 1000 & Number of test episodes \\
        \bottomrule
    \end{tabular}

    \label{tab:parameter_table}
\end{table}

In the CartPole problem, the action space $\mathcal{A}$ contains just two actions $\mathtt{Left}$ and $\mathtt{Right}$, which push the cart to the left or to the right, respectively. At the end of each step, depending on the true state $s_t$, and on the taken action $a_t$, the environment will return a deterministic reward $ R_t = -x_{\text{max}}^{-1} |x_t|-\psi_{\text{max}}^{-1} |\psi_t|, \  \forall t$,
where $x_{\text{max}} = 4.8$~m and $\psi_{\text{max}} = \frac{2\pi}{15}$~rad (equivalent to $24^\circ$) are the maximum values for the two quantities. If the angle or cart position go outside the boundaries, the episode is over, and the agents do not accumulate any more reward. The goal for the two agents is thus to maximize the cumulative discounted sum of the reward $R_t$, while limiting the communication cost.

\subsection{The Coding and Decoding Functions}
\label{subsec:training_vqvae}

\begin{figure}[t]
\centering
\subfloat[PSNR.\label{fig:vqvae}]{\input{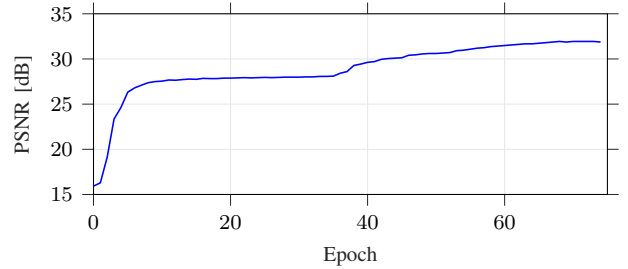}}\\
\subfloat[Perplexity.\label{fig:perplexity}]{\input{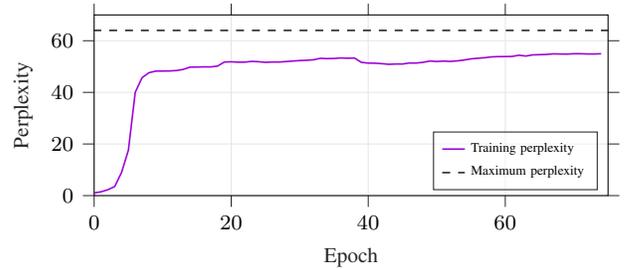}}%
\caption{Training of the \gls{vqvae} model $\zeta_6$, with 6 bits per feature.}
\label{fig:vqvae_training}
\end{figure}

As mentioned in Sec.~\ref{ssec:coding}, the observer can optimize its coding function $\Lambda$ according to different criteria depending on the considered communication problem. However, as we explained in Sec.~\ref{sec:solution}, optimizing $\Lambda$ without any parameters is usually not feasible due to the curse of dimensionality on the action space. Consequently, we rely on a pre-trained set $\mc{V}$ of \gls{vqvae} models, whose codebooks are optimized to solve the technical problem, i.e., minimizing the distortion on the observation measured using the \gls{mse}: $d_{A}(o, \hat{o}) = \text{MSE}(o, \hat{o})$. The training performance of the \gls{vqvae} with $\zeta_6$\edit{, i.e., using the maximum value $V$ of 6 bits per feature,} is shown in Fig.~\ref{fig:vqvae_training}: the encoder converges to a good reconstruction performance, which can be measured by its \emph{perplexity}. The perplexity is simply $2^{H(p)}$, where $H(p)$ is the entropy of the codeword selection frequency, and a perplexity equal to the number of codewords is the theoretical limit, which is only reached if all codewords are selected with the same probability. The perplexity at convergence is 54.97, which is close to the theoretical limit for a real application.

\begin{figure*}[t]
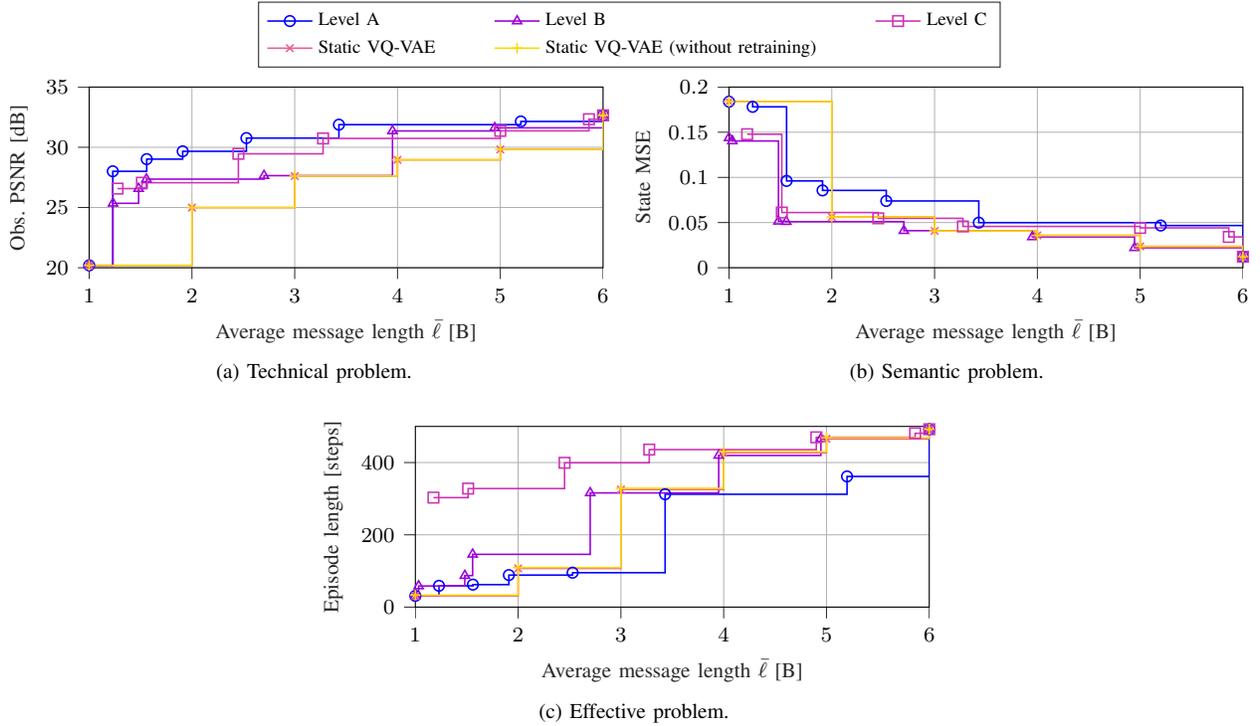

\centering
    \subfloat{\input{figures/prob_legend}}\vspace{-0.3cm}\\
        \setcounter{subfigure}{0}
\subfloat[Technical problem.\label{fig:technical}]{\input{figures/tech_prob}}
\subfloat[Semantic problem.\label{fig:semantic}]{\input{figures/sem_prob}}\\
\subfloat[Effective problem.\label{fig:effective}]{\input{figures/eff_prob}}%
\caption{Performance of the communication schemes on the three Levels of the remote \gls{pomdp}.}
\label{fig:performance}
\end{figure*}

The observer then uses \gls{drl} to foresightedly \edit{choose the best \gls{vqvae}} $\zeta_v$ at each time step, maximizing the expected long-term reward for each communication problem. We train the observer to solve each specific coding problem by designing three different rewards, depending on the considered communication Level \edit{(A, B, or C)}:
\begin{enumerate}
\item \emph{Level A (technical problem):} The distortion metric for the observer is $d_A(o_t,\hat{o}_t)=-\text{PSNR}(o_t, \hat{o_t})$, as part of the reward definition from~\eqref{eq:rew_A}. The \gls{psnr} is an image fidelity measure proportional to the logarithm of the normalized \gls{mse} between the original and reconstructed image;
\item \emph{Level B (semantic problem):} The  distortion metric is $d_B(\hat{s}^{(o)}_t,\hat{s}^{(r)}_t) = \text{MSE}(\hat{s}^{(o)}_t, \hat{s}^{(r)}_t)$, as part of the reward defined in~\eqref{eq:rew_B}, and the decoder needs to estimate the underlying physical state $s_t$ by minimizing the \gls{mse}, i.e., the distance between $\hat{s}^{(o)}_t$ and $\hat{s}^{(r)}_t$ in the semantic space. In our case, the estimator used to obtain the estimates is a pre-trained supervised \gls{lstm} neural network;
\item \emph{Level C (effective problem):} In this case, there is no direct distortion metric, and the control performance is used directly as in~\eqref{eq:rew_C}. The policy $\pi^{(r)}$ is given by an actor-critic agent implementing an \gls{lstm} architecture, pre-trained using data with the highest available message quality (6 bits per feature).
\end{enumerate}
In this case, the task depends on all the semantic features contained in $S_t$. However, the $4$ components of the state do not carry the same amount of information to the robot: depending on the system conditions, i.e., the state $S_t$, some pieces of information are more relevant than others.

\subsection{Neural Network Architecture and Training}
\label{subsec:parameters} 

The \gls{vqvae} architecture is made with \gls{cnn} layers to extract latent features and it is trained separately before the training of the control policy. To this end, a dataset of observations is collected through a random policy. We then train an encoding network, the vector quantization layer and the decoder jointly as in the standard \gls{vqvae} \cite{van2017neural}. The first vector quantization layer learned contains the highest number of codewords. Finally, we fix the encoder and the decoder and just train the other vector quantization layers, obtaining multiple quantizers over the same latent space discovered by a common encoder. The hyperparameters used to train the \gls{vqvae} are reported in Table~\ref{tab:parameter_table}. After obtaining the $V$ quantizers, we train the policy using the standard \gls{a2c} algorithm. Table~\ref{tab:nn_params} shows the Encoder-Decoder layers of the \gls{vqvae}. In Table~\ref{tab:rec_models}, the layers of the implemented Regressor and the Actor-critic neural networks are reported. All the \gls{nn}s are implemented through the Pytorch library.  
Once the robot policy has been obtained, we can train the observer policy. The observer learns a policy through the same \gls{a2c} algorithm, but in this case the input to the policy are the features before quantization. A unique observer policy is trained for different values of the trade-off parameter $\beta$ and for different communication Levels. For further details on the implementation, training and testing process, we refer to the publicly available simulation code.\footnote{\url{https://www.github.com/pietro-talli/tmlcn_code}}

\edit{The additional computational cost of the architecture on the observer side is well within the computational capabilities of even relatively simple embedded devices~\cite{kang2021benchmarking}, and even more complex problems can be dealt with by Edge devices. At the same time, training the actor with compressed representations actually reduces its computational burden, as the feature extraction is performed by the sender. However, if the sender is significantly computationally constrained, replacing the \gls{vqvae} with a classical compression scheme such as JPEG might be a good way to reduce the cost and still deliver the benefits of dynamic compression.}

\begin{table}[t]
    \centering
    \caption{Encoder-Decoder parameters.}
    \begin{tabular}{cccc}
        \toprule
        Layer type & Size & Kernel size & Stride \\
        \midrule
        \multicolumn{4}{c}{Encoder} \\
        $\mathtt{Conv2d + ReLU}$ & $64$ & $10\times 11$ & $8\times 9$ \\ 
        $\mathtt{Conv2d + ReLU}$ & $64$ & $12\times 12$ & $10\times 10$ \\ 
        $\mathtt{Conv2d + ReLU}$ & $128$ & $3\times 3$ & $1\times 1$ \\ 
        $\mathtt{ResidualStack}$ & 2 & $3\times 3$ &  $1\times 1$ \\
        $\mathtt{Conv2d}$ & $8$ & $3\times 3$ & $1\times 1$ \\ 
        \midrule
        \multicolumn{4}{c}{Decoder} \\
        $\mathtt{ResidualStack}$ & 2 & $3\times 3$ &  $1\times 1$ \\
        $\mathtt{Conv2d + ReLU}$ & $128$ & $3\times 3$ & $1\times 1$ \\ 
        $\mathtt{Conv2d + ReLU}$ & $64$ & $12\times 12$ & $10\times 10$ \\ 
        $\mathtt{Conv2d}$ & $64$ & $10\times 11$ & $8\times 9$ \\ 

        \bottomrule
    \end{tabular}
    
    \label{tab:nn_params}
\end{table}
\begin{table}[t]
    \centering
    \caption{Recurrent architectures.}
    \begin{tabular}{cccl}

        \midrule
        Layer type & Inputs & Outputs & Description \\ 
        \midrule
        \multicolumn{4}{c}{Regressor} \\
        $\mathtt{LSTM + ReLU}$ & $64$ & $64$ & Single recurrent layer \\
        $\mathtt{Linear + ReLU}$ & $64$ & $128$ & Hidden layer \\
        $\mathtt{Linear}$ & $128$ & $1$ & Output layer \\
        \midrule
        \multicolumn{4}{c}{Actor-critic} \\
        $\mathtt{LSTM + ReLU}$ & $64$ & $64$ & Single recurrent layer \\
        $\mathtt{Linear + ReLU}$ & $64$ & $128$ & Hidden layer \\
        $\mathtt{Linear}$ & $128$ & $1$ & Output layer (Value) \\
        $\mathtt{Linear + softmax}$ & $128$ & $\lvert \mathcal{A} \rvert$ & Output layer (Policy) \\
        \bottomrule
    \end{tabular}
    
    \label{tab:rec_models}
\end{table}

\subsection{Results}
\label{subsec:results} 
We assess the performance of the three different tasks in the CartPole scenario by simulation, measuring the results over 1000 episodes after convergence. Fig.~\ref{fig:performance} shows the performance of the various schemes over the three problems, compared with a static \gls{vqvae} solution with a constant compression level. In the Level C evaluation, we also consider a static \gls{vqvae} solution in which the robot is not retrained for each $v$, but is only trained for $v=6$ \edit{bits per feature} (i.e., 48 bits per message\edit{, as the \gls{vqvae} considers 8 features}) as for the dynamic scheme. We trained the dynamic schemes with different values of the communication cost $\beta$, so as to provide a full picture of the adaptation to the trade-off between performance and compression. We also introduce the notion of \emph{Pareto dominance}: an $n$-dimensional tuple $\eta=(\eta_1,\ldots,\eta_n)$ Pareto dominates $\eta'$ (which we denote as $\eta\succ \eta'$) if and only if:
\begin{equation}
\eta\succ \eta' \iff \exists i:\eta_i>\eta_i' \wedge\eta_j\geq\eta_j'\,\forall j.
\end{equation}
We can extend this to schemes with multiple possible configurations. The definition of Pareto dominance for schemes $x$ and $y$ is: $x\succ y \iff \exists \eta(x)\succ\eta(y)\,\forall \eta(y)$, i.e., for each configuration of scheme $y$, there is a setting of $x$ that Pareto dominates it. In other words, we can always tune scheme $x$ so that it outperforms any configuration of scheme $y$ on all metrics.

\begin{figure*}[t]
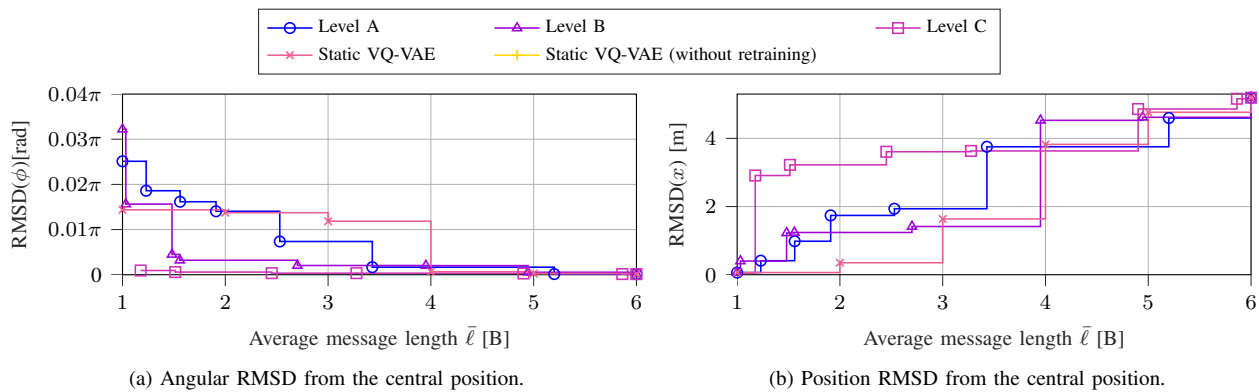

\centering
    \subfloat{\input{figures/prob_legend}}\vspace{-0.3cm}\\
        \setcounter{subfigure}{0}
\subfloat[Angular RMSD from the central position.\label{fig:angular_RMSD}]{\input{figures/RMS_angle}}
\subfloat[Position RMSD from the central position.\label{fig:position_RMSD}]{\input{figures/RMS_position}}%
\caption{Other performance metrics relative to the CartPole control problem.}
\label{fig:RMS_angle_and_pos}
\end{figure*}

We first consider the technical problem performance, shown in Fig.~\ref{fig:technical}: as expected, the Level A dynamic compression outperforms all other solutions, and its performance is Pareto dominant with respect to static compression. Interestingly, the Level B and Level C solutions perform worse than static compression: by concentrating on features in the semantic space or the task space, these solutions remove information that could be useful to reconstruct the full observation, but is meaningless for the specified task.

In the semantic problem, shown in Fig.~\ref{fig:semantic}, a lower \gls{mse} on the reconstructed state is better, and the Level B solution is Pareto dominant with respect to all others. The Level A solution also Pareto dominates static compression, while the Level C solution only outperforms it for higher compression levels, i.e., on the left side of the graph.

Finally, Fig.~\ref{fig:effective} shows the performance of the effectiveness problem (Level C), summarized by how long the CartPole system manages to remain within the position and angle limits. The Level C solution significantly outperforms all others, but is not strictly Pareto dominant: when the communication constraint is very tight, setting a static compression and retraining the robot to deal with the specific \gls{vqvae} used may provide a slight performance advantage. In general, almost perfect control can be achieved with less than half of the average bitrate of the static compressor, which can only reach a similar performance at a much higher communication cost. We also note that, in this case, the Level B solution performs worst: choosing the solution that minimizes the semantic distortion is not always matched to the task, \edit{as it considers the state variables as having equal weight}, while a higher precision may be required when the quantization error affects the robot's action.

Another analysis is conducted on the way the CartPole is controlled with the different communication policies. Fig.~\ref{fig:RMS_angle_and_pos} shows the Angular \gls{rmsd} (Fig.~\ref{fig:angular_RMSD}) and the Position \gls{rmsd} (Fig.~\ref{fig:position_RMSD}), defined as: 
\begin{equation*}
    \text{RMSD}(x) = \sqrt{\frac{1}{t}  \sum_{i = 1}^t (x_i - x_{\text{target}})^2},
\end{equation*}
where $x_{\text{target}}$ is the desired value of the controlled process and $x_i$ is the recorded value of the process at time step $i$. 
Both \gls{rmsd} are computed with respect to the central and vertical position of the CartPole: $x_{\text{target}} = 0$ and $\psi_{\text{target}} = 0$. These results help to evaluate how well the control dynamics keep the CartPole near the optimal central position and to assess the smoothness of the resulting pole oscillations. It is possible to see that, in general, a higher rate allows to keep the angular \gls{rmsd} smaller. In particular, in the Level C system, the values are the smallest. However, this comes at the cost of deviating more from the central position, as shown in the figure. The policy prioritizes the stabilization of the pole oscillations, though this requires deviating from the central position. 
This is because swings in the pole's angle are harder to control due to the instability of the inverted pendulum, and there is a significant risk that the pole might go out of the acceptable range, ending the episode.
\begin{figure}
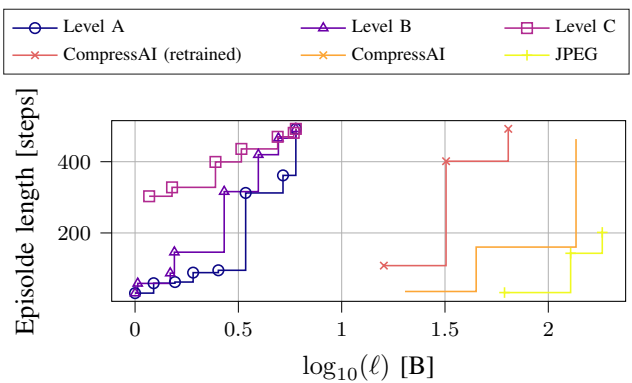

    \centering
    \subfloat{\input{figures/legend_comparison}}\vspace{-0.3cm}\\
        \setcounter{subfigure}{0}
\subfloat{\input{figures/comparison_plot}}
    \caption{Comparison with other methods.}
    \label{fig:comparison-plot}
\end{figure}

\begin{figure*}[t]
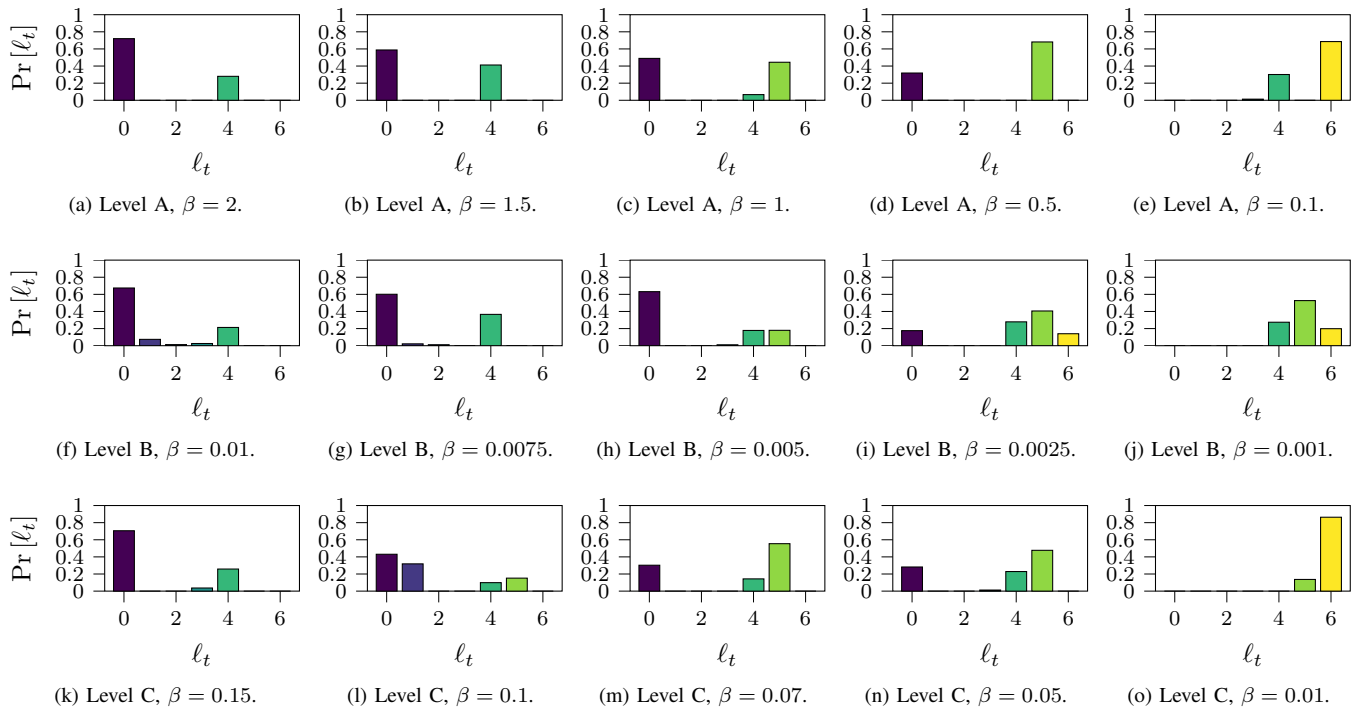

\centering
\subfloat[Level A, $\beta = 2$.\label{fig:level_A_dist_2}]{\input{figures/k_distributions/level_A_distribution_2}}
\subfloat[Level A, $\beta = 1.5$.\label{fig:level_A_dist_15}]{\input{figures/k_distributions/level_A_distribution_1.5}}
\subfloat[Level A, $\beta = 1$.\label{fig:level_A_dist_1}]{\input{figures/k_distributions/level_A_distribution_1}}
\subfloat[Level A, $\beta = 0.5$.\label{fig:level_A_dist_05}]{\input{figures/k_distributions/level_A_distribution_0.5}}
\subfloat[Level A, $\beta = 0.1$.\label{fig:level_A_dist_01}]{\input{figures/k_distributions/level_A_distribution_0.1}}\\
\subfloat[Level B, $\beta = 0.01$.\label{fig:level_B_dist_01}]{\input{figures/k_distributions/level_B_distribution_0.01}}
\subfloat[Level B, $\beta = 0.0075$.\label{fig:level_B_dist_0075}]{\input{figures/k_distributions/level_B_distribution_0.0075}}
\subfloat[Level B, $\beta = 0.005$.\label{fig:level_B_dist_005}]{\input{figures/k_distributions/level_B_distribution_0.005}}
\subfloat[Level B, $\beta = 0.0025$.\label{fig:level_B_dist_0025}]{\input{figures/k_distributions/level_B_distribution_0.0025}}
\subfloat[Level B, $\beta = 0.001$.\label{fig:level_B_dist_001}]{\input{figures/k_distributions/level_B_distribution_0.001}}\\
\subfloat[Level C, $\beta = 0.15$.\label{fig:level_C_dist_015}]{\input{figures/k_distributions/level_C_distribution_0.15}}
\subfloat[Level C, $\beta = 0.1$.\label{fig:level_C_dist_01}]{\input{figures/k_distributions/level_C_distribution_0.1}}
\subfloat[Level C, $\beta = 0.07$.\label{fig:level_C_dist_007}]{\input{figures/k_distributions/level_C_distribution_0.07}}
\subfloat[Level C, $\beta = 0.05$.\label{fig:level_C_dist_005}]{\input{figures/k_distributions/level_C_distribution_0.05}}
\subfloat[Level C, $\beta = 0.01$.\label{fig:level_C_dist_001}]{\input{figures/k_distributions/level_C_distribution_0.01}}
\caption{Distribution of the selected compression levels.}
\label{fig:compression_dist}
\end{figure*}

\subsection{Comparison with existing compression approaches}
\label{subsec:comparison}
\edit{In this section, we compare the performance of our proposed approach to that of other methods. Specifically, we show how digital compression techniques, such as JPEG, perform in the same scenario. We also compare other \gls{nn}-based compression models and evaluate their performance. For the digital compression, we use different sets of parameters combining image resizing, the quality parameter of the JPEG standard, and the number of color grayscale levels. For learning-based compression, we used the \texttt{CompressAI} library which implements the model proposed in \cite{cheng2020image}}.

\edit{Fig.~\ref{fig:comparison-plot} shows the performance of other methods with respect to the proposed dynamic feature compression method. It is possible to see that the digital compression scheme does not allow the actor to effectively control the system, as its stability is low even though the updates are bigger by two orders of magnitude. Obtaining a high control performance would require an extremely high bitrate. On the other hand, the neural compression scheme achieves a higher performance with a limited bitrate, but since the model is a general compression technique designed to compress a wide variety of images, it cannot reach extremely low bitrates. After retraining the scheme on CartPole pictures, it is possible to obtain lower bitrates while improving the resulting control performance. Our approach, which is directly trained on the CartPole task, outperforms all others; however, we do not claim that \gls{vqvae} is the best compression technique for all scenarios. The main contribution of this work is not in the specific compression scheme, but rather in the dynamic and goal-oriented adaptation of the compression parameters for each transmitted update, which could be directly applied to different \gls{nn} architectures and even JPEG.}

\subsection{Analysis of the communication policy}

We can then use an explainability approach to gain further insights on how effective communication operates. Fig.~\ref{fig:compression_dist} shows the distribution of the quantization level selected by an observer trained for each communication Level (A, B, and C). We note that \edit{we adapted the scale of $\beta$ for the three Levels, so as to obtain comparable results: as the reward process takes values in different ranges (e.g., the \gls{psnr} is in dB while the reward of the \gls{mdp} is between $-1$ and $1$), using the same transmission cost would result in very different outcomes. We then chose to rescale the transmission cost parameter to have the full range of average bitrates at each communication Level.} The similarity in the compression level distributions at the three Levels is striking. For lower values of $\beta$, the observer selects $\zeta_\varnothing$, which corresponds to no transmission, more often. As $\beta$ decreases, the communication cost becomes lower, and thus the observer chooses longer messages more often. Another common pattern is that quantizing features using $1,2$ or $3$ bits is a rare choice. This shows that the memory implemented implicitly in the system through the \gls{lstm} is powerful enough to obtain adequate beliefs based on past messages, so that the observer can rely on it and not send anything. \edit{A roughly quantized update has a relatively low value, as its novelty is limited, and transmitting intermittent updates at a higher quality results in a better performance.}

\begin{figure*}[t]
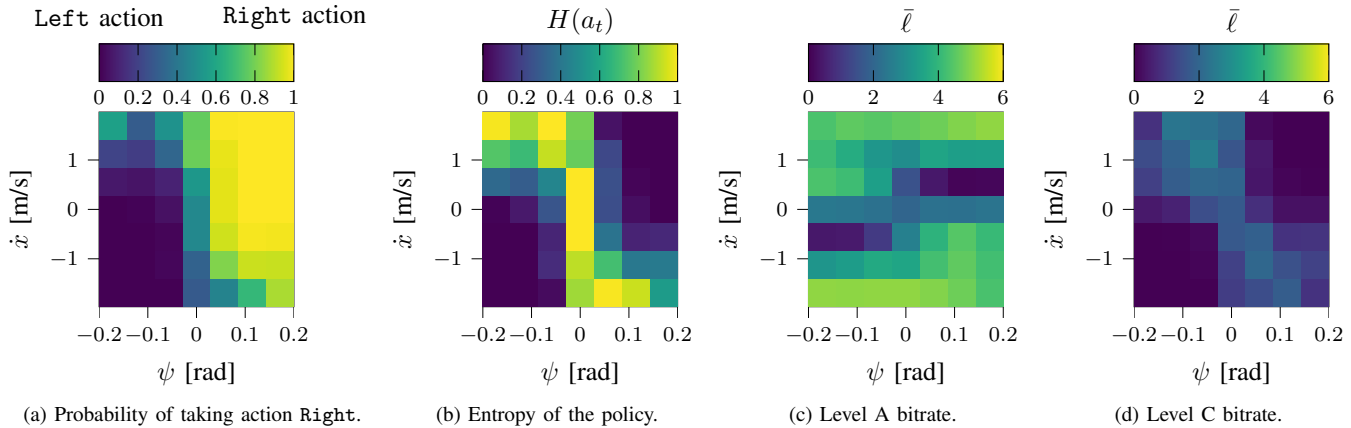

\centering
\setcounter{subfigure}{0}
\subfloat[Probability of taking action $\mathtt{Right}$.\label{fig:control_actions_pos_vs_angle}]{\input{figures/control_actions_pos_vs_angle}}%
\subfloat[Entropy of the policy.\label{fig:policy_entropy_pos_vs_angle}]{\input{figures/policy_entropy_pos_vs_angle}}%
\subfloat[Level A bitrate.\label{fig:bits_per_feature_pos_angle_A}]{\input{figures/bits_per_feature_pos_vs_angle_A}}%
\subfloat[Level C bitrate.\label{fig:bits_per_feature_pos_angle_C}]{\input{figures/bits_per_feature_pos_vs_angle}}%
\caption{Analysis of the transmission policy as a function of the pole angle $\psi$ and cart linear velocity $\dot{x}$. \edit{The bitrate is measured in Bytes per transmission.}}
\label{fig:pos_vs_angle}
\end{figure*}

\begin{figure*}[t]
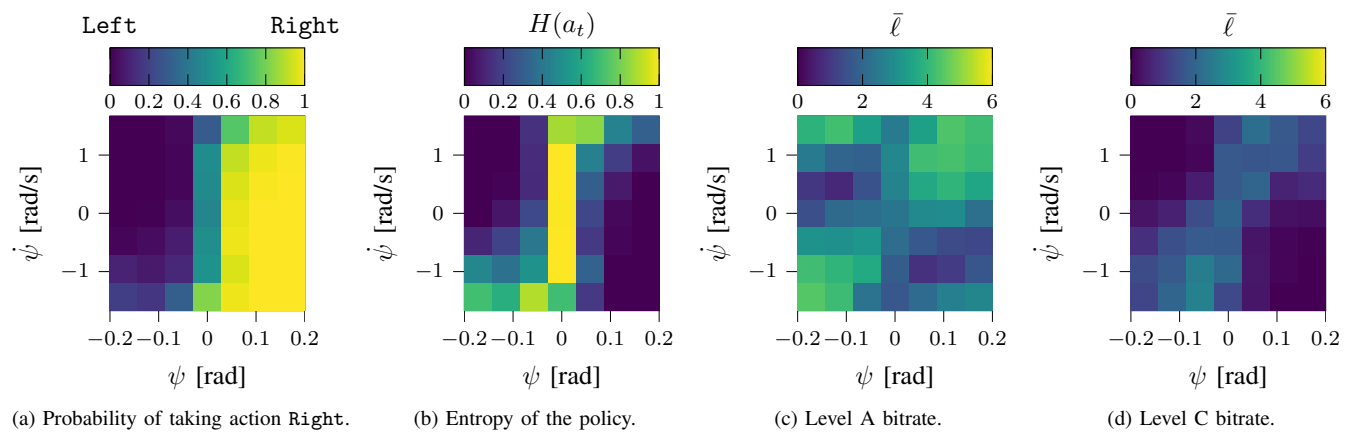

\centering
\setcounter{subfigure}{0}
\subfloat[Probability of taking action $\mathtt{Right}$.\label{fig:control_actions_omega_vs_angle}]{\input{figures/control_actions_omega_vs_angle}}%
\subfloat[Entropy of the policy.\label{fig:policy_entropy_omega_vs_angle}]{\input{figures/policy_entropy_omega_vs_angle}}%
\subfloat[Level A bitrate.\label{fig:bits_per_feature_omega_angle_A}]{\input{figures/bits_per_feature_omega_vs_angle_A}}%
\subfloat[Level C bitrate.\label{fig:bits_per_feature_omega_angle_C}]{\input{figures/bits_per_feature_omega_vs_angle}}%
\caption{Analysis of the transmission policy as a function of the pole angle $\psi$ and angular velocity $\dot{\psi}$. \edit{The bitrate is measured in Bytes per transmission.}}
\label{fig:omega_vs_angle}
\end{figure*}

However, the real difference between the three policies is given by \emph{when} they decide not to transmit. Therefore, we propose an analysis based on the visualization of the observer policy and the receiver policy. In Fig.~\ref{fig:pos_vs_angle}, four colormaps show different policies projected in the same domain: the pole angle on the x-axis and the cart velocity on the y-axis. More specifically, we quantize the projected state into cells and show the policy of the robot and the observer in each cell. Fig.~\ref{fig:control_actions_pos_vs_angle} shows the robot actions, averaged among $10^6$ samples. Since in the CartPole problem the actions are binary, we represent the probability of choosing action $\mathtt{Right}$ in a range between $0$ and $1$. Fig.~\ref{fig:policy_entropy_pos_vs_angle} depicts the entropy of the robot policy. We considered the action probability in the previous figure and compute the action entropy as follows:
\begin{equation*}
H(a) = - p(a\!=\!0) \log_2\left(p(a\!=\!0)\right)-p(a\!=\!1) \log_2\left(p(a\!=\!1)\right),
\end{equation*}
where $p(a)$ is empirically estimated by counting the number of times each action is chosen when the state is in the projected cell. 
Fig.~\ref{fig:bits_per_feature_pos_angle_A} and Fig.~\ref{fig:bits_per_feature_pos_angle_C} show the average \edit{length of the update packets, i.e., the} number of \edit{Bytes} transmitted in each cell when optimizing for Levels A and C, respectively. This can be seen as the average number of bits that the transmitter allocates for each projected slice of the state space. Fig.~\ref{fig:omega_vs_angle} shows the same results but for a different physical state projection, mapping the angle $\psi$ on the x-axis and the pole angular velocity $\dot{\psi}$ on the y-axis.

In both figures, there is a strong correspondence between the states where the robot entropy is higher and the states where the Level C policy allocates a higher number of bits. This confirms that an effective observer policy manages to discriminate the uncertainty at the robot side. In regions of the state space where it is more difficult to retrieve the correct action, i.e., the action entropy is higher, the observer will provide the robot with more precise information by sending longer messages. There are regions where the robot action is always the same, e.g., whenever the cart is moving fast and the tip of the pole is pointing to the same side the cart is moving towards. In these cases, the entropy is extremely low, and the transmitter can avoid sending new updates to the robot. This is due to the fact that, even if the estimated state at the receiver differs from the observed one, the action to perform remains the same and will be to push further the cart to try to get the pole more vertical. Recalling~\eqref{eq:voi_C}, we note that if $Q\left(\xi^{(r)}_t,\pi^{(r)}\!\left(\xi^{(r)}_t\right)\!\right)$ is very sensitive to small variations in $\xi^{(r)}_t$, then the gap in~\eqref{eq:voi_C} is going to be significant, leading the observer to choose to send precise information. In principle, a Level C transmitter could reduce the message length or even avoid transmission as long as the robot is able to choose the correct actions, even though its belief is incorrect. An optimal communication scheme approximately follows
\begin{equation*}
    \ell_t \propto V\!\left(\xi^{\text{pri},(r)}_t,m\!\right) ,
\end{equation*}
which means that the message length is roughly proportional to \gls{voi}. This concept might be used when defining a heuristic policy, which behaves similarly to the effective communication policy but is much simpler to design and implement. Note that this condition includes two separate cases in which a Level C observer chooses not to transmit, while Level A and B transmitters would send precise data: \begin{itemize}
    \item The action corresponding to the prior belief is the same as the one after the updating message. In this case, the \gls{voi} of the communicated message is low and thus we can lower $\ell_t$;
    \item The action is different after the communicated message, but the long-term rewards are close enough that the robot is not going to benefit too much from choosing the other action. Even in this case, sending less information is not going to affect the control performance significantly.
\end{itemize}
These cases cannot be taken into account in Levels A and B. Indeed, the Level A policy shown in Fig.~\ref{fig:bits_per_feature_pos_angle_A} tends to allocate communication resources in the states where the picture is changing more rapidly, so that the memory available to the robot is less useful to estimate the current observation, regardless of the correct action. As the cart speed $\dot{x}$ increases along the y-axis, the number of bits increases too. The same reasoning can be applied to the results in Fig.~\ref{fig:omega_vs_angle}. 

Another general principle that we can deduce for an effective policy is that it should be aware of variations of the value function with respect to the belief. If the value function is strongly affected by small perturbations of the belief, then the effective policy should communicate more information in order to reduce the discrepancy between $\xi^{(o)}_t$ and $\xi^{(r)}_t$. This reasoning can be intuitively understood by looking at the differential of the robot's value function $Q\left(\xi^{(r)}_t,\pi^{(r)}\!\left(\xi^{(r)}_t\right)\!\right)$ with respect to changes in its belief distribution $\xi^{(r)}_t$.
When this value is big, an inaccurate estimation of the state would cause a poor estimation of the value function, which may in turn cause the robot to choose a low-quality action. 
\begin{figure*}[t]
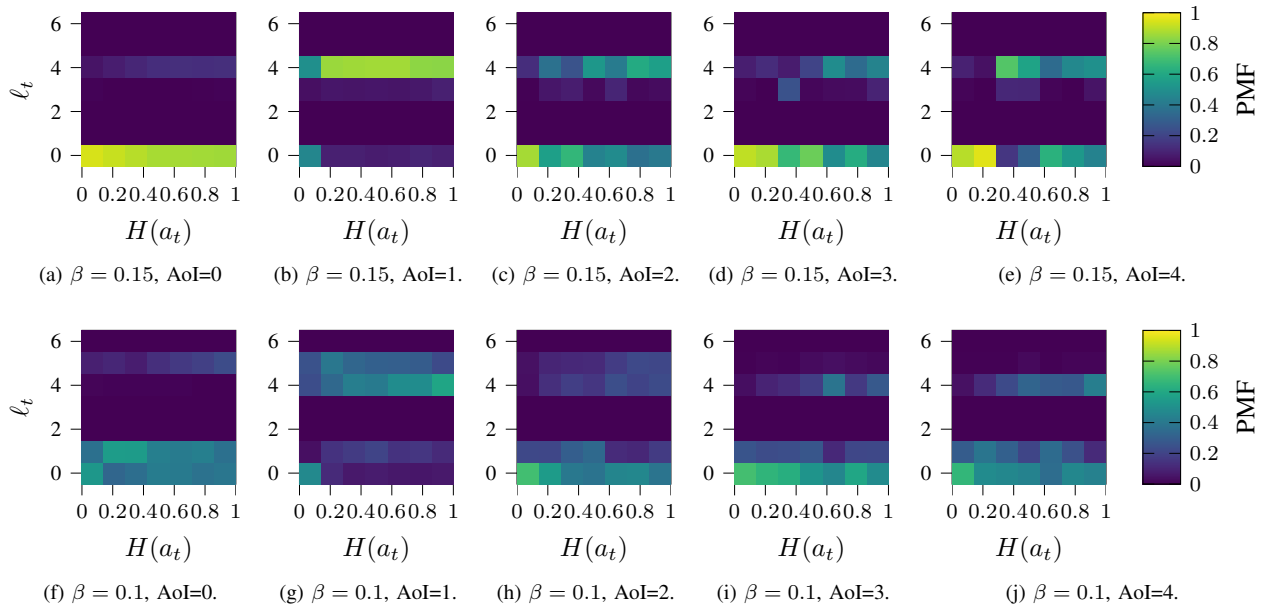

\centering
\subfloat[$\beta=0.15$, \gls{aoi}=0\label{fig:H_vs_dist_015_0}]{\input{figures/H_vs_dist/entropy_vs_sensor_action_0150}}%
\subfloat[$\beta=0.15$, \gls{aoi}=1.\label{fig:H_vs_dist_015_1}]{\input{figures/H_vs_dist/entropy_vs_sensor_action_0151}}%
\subfloat[$\beta=0.15$, \gls{aoi}=2.\label{fig:H_vs_dist_015_2}]{\input{figures/H_vs_dist/entropy_vs_sensor_action_0152}}%
\subfloat[$\beta=0.15$, \gls{aoi}=3.\label{fig:H_vs_dist_015_3}]{\input{figures/H_vs_dist/entropy_vs_sensor_action_0153}}
\subfloat[$\beta=0.15$, \gls{aoi}=4.\label{fig:H_vs_dist_015_4}]{\input{figures/H_vs_dist/entropy_vs_sensor_action_0154}}\\
\subfloat[$\beta=0.1$, \gls{aoi}=0.\label{fig:H_vs_dist_01_0}]{\input{figures/H_vs_dist/entropy_vs_sensor_action_010}}%
\subfloat[$\beta=0.1$, \gls{aoi}=1.\label{fig:H_vs_dist_01_1}]{\input{figures/H_vs_dist/entropy_vs_sensor_action_011}}%
\subfloat[$\beta=0.1$, \gls{aoi}=2.\label{fig:H_vs_dist_01_2}]{\input{figures/H_vs_dist/entropy_vs_sensor_action_012}}%
\subfloat[$\beta=0.1$, \gls{aoi}=3.\label{fig:H_vs_dist_01_3}]{\input{figures/H_vs_dist/entropy_vs_sensor_action_013}}%
\subfloat[$\beta=0.1$, \gls{aoi}=4.\label{fig:H_vs_dist_01_4}]{\input{figures/H_vs_dist/entropy_vs_sensor_action_014}}%
\caption{Level C observer action distribution for different robot action entropy levels and values of $\beta$. \edit{The color of each cell represents the empirical probability of choosing a packet length $\ell_t$ for a given entropy $H(a_t)$ and \gls{aoi}. Each column of each subfigure then represents a conditional \gls{pmf}.}}\vspace{-0.4cm}
\label{fig:H_vs_dist_C}
\end{figure*}

\begin{figure*}[t]
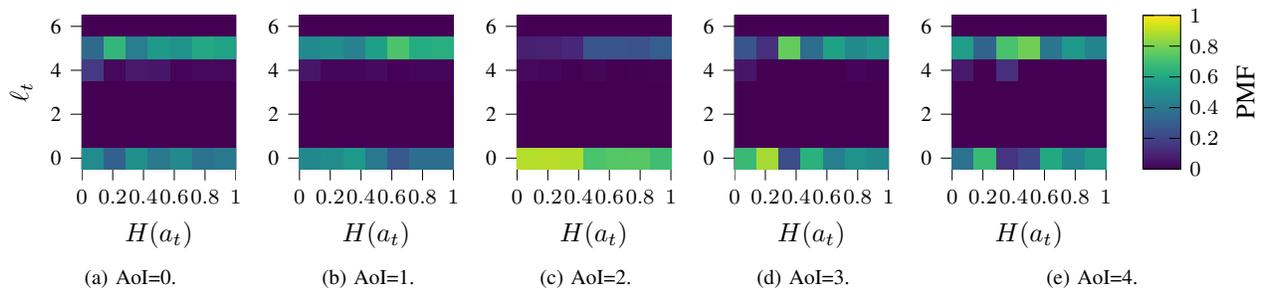

\centering
\setcounter{subfigure}{0}
\subfloat[\gls{aoi}=0.\label{fig:H_vs_dist_A_0}]{\input{figures/H_vs_dist/entropy_vs_sensor_action_A_150}}%
\subfloat[\gls{aoi}=1.\label{fig:H_vs_dist_A_1}]{\input{figures/H_vs_dist/entropy_vs_sensor_action_A_151}}%
\subfloat[\gls{aoi}=2.\label{fig:H_vs_dist_A_2}]{\input{figures/H_vs_dist/entropy_vs_sensor_action_A_152}}%
\subfloat[\gls{aoi}=3.\label{fig:H_vs_dist_A_3}]{\input{figures/H_vs_dist/entropy_vs_sensor_action_A_153}}%
\subfloat[\gls{aoi}=4.\label{fig:H_vs_dist_A_4}]{\input{figures/H_vs_dist/entropy_vs_sensor_action_A_154}}%

\caption{Level A observer action distribution for different robot action entropy levels with $\beta=1$.}
\label{fig:H_vs_dist_A}
\end{figure*}

In Fig.~\ref{fig:H_vs_dist_C}, we provide an analysis of the communication strategy with respect to different \gls{aoi} values. This allows to show how the memory of the robot and of the observer plays a crucial role on the communication decisions. In particular, we consider five values of the \gls{aoi}: \gls{aoi} $=0$ indicates that a message of any length was transmitted in the previous time step. \gls{aoi} $=\Delta$ with $\Delta \in \lbrace 1,2,3,4 \rbrace$ means that no messages have been received by the observer for $\Delta$ time steps since the last received message. This is a measure of how up to date the memory of the robot is, allowing us to evaluate the next choice of the observer for a given age. We then consider the distribution of the observer actions (y-axis) with respect to different ranges of the robot actions entropy (x-axis). This means that, for each entropy interval, we count the number of times each action is performed, in order to obtain an empirical distribution. The columns are normalized so that each cell shows the probability that the observer chooses a specific $\ell_t$ whenever the robot action entropy falls within the corresponding interval, for different values of the \gls{aoi}. 

Fig.~\ref{fig:H_vs_dist_015_0} clearly shows that, if there was a transmission in the previous time step (\gls{aoi} $=0$), it is very unlikely that the system is going to be updated again in the current time step. \edit{As remarked above, Figs.~\ref{fig:H_vs_dist_015_0}-\subref*{fig:H_vs_dist_015_4} show the case with $\beta=0.15$, for which the observer almost always picks $\zeta_4$ to transmit. On the other hand, Figs.~\ref{fig:H_vs_dist_01_0}-\subref*{fig:H_vs_dist_01_4} show the case with $\beta=0.1$, in which the agent sometimes selects other codebooks due to the lower transmission costs.} 
The observer often chooses to communicate if \gls{aoi} $=1$, with an exception if the system is in a very low entropy state, in which case the probability of communicating using $\zeta_4$ is similar to the one corresponding to action $\zeta_{\varnothing}$. If we look at the behavior for higher values of the \gls{aoi}, we can notice a general trend: communication is more likely to happen in higher entropy states than in lower entropy ones. This shows that the observer policy understands the cases where the state has to be precisely estimated by the robot to choose its action correctly. \edit{Additionally, the probability of transmitting an update actually \emph{decreases} when the \gls{aoi} reaches higher values. The observer will only skip several consecutive transmission opportunities in two cases: either the system state is highly predictable, and the actor can rely on its past knowledge to get a precise estimate of it, or the two actions are almost equivalent, e.g., if the pole is balanced vertically. The former case is more likely to be a low-entropy state, while the latter is high-entropy, but has a small difference between the rewards for the two actions.}
We can see that the trend holds for different values of $\beta$ by looking at Figs.~\ref{fig:H_vs_dist_01_0}-\subref*{fig:H_vs_dist_01_4}: if we decrease the value of $\beta$, the observer tends to transmit more often, and use higher message lengths when it transmits, but the general tendency to transmit more whenever the robot action entropy is high clearly holds. This final analysis allows us to get an easy heuristic for effective communication when the value function is not available or cannot be learned.

Fig.~\ref{fig:H_vs_dist_A} shows that the Level A policy allocates communication resources without considering the entropy of the control actions. \edit{While Fig.~\ref{fig:H_vs_dist_C} showed a clear monotonic trend in the probability of selecting $\zeta_4$, which increased as the entropy increased, the pattern is much weaker in this case. } The value of $\beta$ for this was chosen to get a similar overall bitrate (and, as we discussed, a similar overall action distribution) to the Level C case with $\beta=0.15$. \edit{As we discussed above, there is a weak correlation between the action entropy and features such as the angular and cart velocities, but it is the latter that the Level A policy considers: as the difficulty of accurately reconstructing the image increases with the speed of the CartPole system, more unstable states have more frequent transmissions.}

\section{Conclusion}
\label{sec:conclusion}

In this work, we presented a dynamic feature compression scheme that can exploit an ensemble \gls{vqvae} to solve the semantic and effective communication problems. The dynamic scheme outperforms fixed quantization, and can be trained automatically with limited feedback, unlike emergent communication models that are unable to deal with complex tasks. The choices made by the observer are clearly tied to the control policy of the robot it aims to help, significantly outperforming a simpler optimization that does not take into account the semantic and effective problems. We also analyzed the optimal policies to draw insights on their decisions, showing that the Level C optimization indeed considers the robot's policy.

A natural extension of this model is to consider more complex tasks and wider communication channels, corresponding to realistic control scenarios, or scenarios with multiple transmitters with partial information about each other and the robot. \edit{Considering a more realistic channel model, which has a loss probability and time-varying statistics in addition to the transmission cost, would also be an interesting direction, joining our model with \gls{jscc} theory. Dynamically adapting \gls{jscc} parameters with the goal of helping a remote \gls{drl} agent would be a natural extension of our proposed approach for more realistic wireless scenarios.}
Another interesting direction for future work is to consider joint training of the robot and the observer, or cases with partial information available at both transmitter and receiver.

\bibliographystyle{IEEEtran}
\bibliography{biblio.bib}

\appendix
\section*{The Information Bottleneck View}
\label{app:info_bottleneck}
We can also consider another perspective on the observer's choices, using information bottleneck theory. We define a \emph{sufficient statistic} $i(s)$ of any given state $s\in\mc{S}$, which is enough to determine the robot's performance in that state. Denoting the number of bits required to represent a realization of random variable $X$ as $b(X)$, we consider a case in which:
\begin{align*}
    b(i(S)) < b(S) < b(O).
\end{align*}
Indeed, the observation may contain much more information than needed to estimate the state~\cite{cover:IT}, and lossily compressing the message to preserve the relevant information, removing redundant or irrelevant details, can ease communication requirements without any performance loss. We can also observe that ${i(S) \rightarrow S \rightarrow O}$ is a Markov chain.
The random quantity $i(S)$ represents the minimal description of the system with respect to the robot's task, i.e., no additional data computed from $S$ adds meaningful information for the robot's policy. The state $S$ may also include task-irrelevant physical information on the system. However, both $S$ and $i(S)$ are unknown quantities, as the observer only receives a noisy and high-dimensional representation of $S$ through $O$. This is a well-known issue in \gls{drl}: in the original paper presenting the \gls{dqn} architecture~\cite{mnih2015human}, the agent could only observe the screen while playing classic arcade videogames, and did not have access to the much more compact and precise internal state representation of the game. Introducing communication and dynamic encoding adds another layer of complexity.

We can then consider the case in which communication is limited to a maximum length of $L$ bits, i.e., to $2^{L+1}-1$ messages, considering all possible lengths lower than or equal to $L$, including no communication. \edit{Naturally, this assumes that the receiver has a way to discriminate between messages of different lengths, e.g., through a MAC layer header.} The channel is ideal, i.e., instantaneous and error-free, but it includes a constant cost per bit $\beta$, as in the observer reward we gave in the previous section. Consequently, the problem introduces an information bottleneck between the observation $O_t$ and the estimate $\hat{o}_t$ that the robot can make, based on the message $M_t$ conveyed through the channel. If we define a distortion measure over the observation space $d_A:\mc{O}^2\rightarrow\mathbb{R}^+$, any communication introduces a non-zero distortion $d_A(o, \hat{o})$ whenever $b(o)>L$, whose theoretical asymptotic limits are given by rate-distortion theory~\cite{cover:IT}.  If we also consider \emph{memory}, i.e., the use of past messages in the estimation of $\hat{o}$, the mutual information between $o$ and the previous messages can be used to reduce the distortion, improving the quality of the estimate.

In the semantic problem, the aim is to extrapolate the real physical state of the system $S_t$ from the compressed observation $M_t$, which can be a complex stochastic function. In general, the real state lies in a low-dimensional semantic space $\mc{S}$. The term semantic is motivated by fact that, in this case, the observer is not just transmitting pure sensory data, but some meaningful piece of physical information about the system. Consequently, the distortion to be considered in this case can be represented by a measure $d_B:\mc{S}^2\rightarrow\mathbb{R}$ over the semantic space, so that the distortion $d_B(\hat{s}^{(o)}_t, \hat{s}^{(r)}_t)$ is computed between the observer's best estimate of the state and the one performed by the robot based on $M_t$, and on its memory of past messages. Finally, to be even more efficient and specific with respect to the task, the observer may optimize the message $M_t$ to minimize a distortion measure
$d_C\left(i\left(\xi^{(o)}_t\right),i\left(\xi^{(r)}_t\right)\right)$ between the effective representation of the observer's belief on the state, which contains only the task-specific information, and the knowledge available to the robot. Naturally, any message instance $m_t\in\mc{M}$ must be at most $L$ bits long, in order to respect the constraint. However, defining the sufficient statistic $i\left(\xi_t^{(o)}\right)$ may be highly complex and problem-dependent, and using the robot's reward as a direct performance measure is significantly more direct, with the same guarantees.





\end{document}

%% file: acronyms.tex
\newacronym{iot}{IoT}{Internet of Things}
\newacronym{ml}{ML}{machine learning}
\newacronym{dl}{DL}{Deep Learning}
\newacronym{marl}{MARL}{multi-agent reinforcement learning}
\newacronym{rl}{RL}{reinforcement learning}
\newacronym{decpomdp}{Dec-POMDP}{Decentralized Partially Observable Markov Decision Process }
\newacronym{pomdp}{POMDP}{Partially Observable Markov Decision Process}
\newacronym{uav}{UAV}{Unmanned Aerial Vehicle}
\newacronym{dqn}{DQN}{Deep Q-Network}
\newacronym{dnn}{DNN}{deep neural network}
\newacronym{dial}{DIAL}{Differentiable Inter-Agent Learning}
\newacronym{mdp}{MDP}{Markov decision process}
\newacronym{fov}{FoV}{Field of View}
\newacronym{cnn}{CNN}{Convolutional Neural Network}
\newacronym{nn}{NN}{neural network}
\newacronym{ddql}{DDQL}{Distributed Deep Q-Learning}
\newacronym{pdf}{PDF}{Probability Density Function}
\newacronym{ndpomdp}{ND-POMDP}{Networked Distributed Partially Observable Markov Decision Process}
\newacronym{radam}{RAdam}{Rectified Adam}
\newacronym{cdf}{CDF}{cumulative distribution function}
\newacronym{mpc}{MPC}{Model Predictive Control}
\newacronym{rv}{rv}{Random Variable}
\newacronym{qoe}{QoE}{Quality of Experience}
\newacronym{tlc}{TLC}{Telecommunications}
\newacronym{cml}{CML}{communications for machine learning}
\newacronym{mlc}{MLC}{machine learning for communications}
\newacronym{drl}{DRL}{Deep Reinforcement Learning}
\newacronym{rf}{RF}{Radio Frequency}
\newacronym{urllc}{URLLC}{Ultra-Reliable and Low-Latency Communications}
\newacronym{fl}{FL}{federated learning}
\newacronym{kpi}{KPI}{Key Performance Indicators}
\newacronym{mec}{MEC}{Mobile Edge Computing}
\newacronym{ei}{EI}{Edge Intelligence}
\newacronym{bs}{BS}{base station}
\newacronym{sdn}{SDN}{Software Defined Networking}
\newacronym{mimo}{MIMO}{Multiple-Input Multiple-Output}
\newacronym{gp}{GP}{Gaussian Process}
\newacronym{iiot}{IIoT}{Industrial Internet of Things}
\newacronym{csi}{CSI}{Channel State Information}
\newacronym{sgd}{SGD}{Stochastic Gradient Descent}
\newacronym{iid}{i.i.d.}{independent and identically distributed}
\newacronym{ofdm}{OFDM}{Orthogonal Frequency Division Multiplexing}
\newacronym{los}{LOS}{Line-of-Sight}
\newacronym{nlos}{NLOS}{Non-Line-of-Sight}
\newacronym{snr}{SNR}{Signal to Noise Ratio}
\newacronym{rb}{RB}{Resource Block}
\newacronym{6g}{6G}{sixth generation}
\newacronym{ai}{AI}{artificial intelligence}
\newacronym{sfl}{SFL}{Synchronous Federated Learning}
\newacronym{frfl}{FRFL}{Fixed Rate Federated Learning}
\newacronym{pgm}{PGM}{Probabilistic Graphical Model}
\newacronym{hmm}{HMM}{Hidden Markov Model}
\newacronym{elbo}{ELBO}{Evidence Lower Bound}
\newacronym{pmf}{PMF}{Probability Mass Function}
\newacronym{smab}{SMAB}{Stochastic Multi-Armed Bandit}
\newacronym{mab}{MAB}{multi-armed bandit}
\newacronym{mc}{MC}{Monte Carlo}
\newacronym{is}{IS}{Importance Sampling}
\newacronym{dms}{DMS}{discrete memoryless source}
\newacronym{ucb}{UCB}{upper confidence bound}
\newacronym{ser}{SER}{Symbol Error Rate}
\newacronym{sc}{SC}{Semantic Communications}
\newacronym{voi}{VoI}{Value of Information}
\newacronym{nlp}{NLP}{natural language processing}
\newacronym{ts}{TS}{Thompson Sampling}
\newacronym{cmab}{CMAB}{contextual multi-armed bandit}
\newacronym{rccmab}{RC-CMAB}{rate-constrained \gls{cmab}}
\newacronym{rcmab}{R-CMAB}{remote \gls{cmab}}
\newacronym{ib}{IB}{information bottleneck}
\newacronym{merl}{MERL}{maximum entropy reinforcement learning}
\newacronym{pac}{PAC}{Probably Approximately Correct}
\newacronym{aoi}{AoI}{Age of Information}
\newacronym{aoii}{AoII}{Age of Incorrect Information}
\newacronym{uoi}{UoI}{Urgency of Information}
\newacronym{vqvae}{VQ-VAE}{Vector Quantized Variational Autoencoder}
\newacronym{vae}{VAE}{Variational Autoencoder}
\newacronym{psnr}{PSNR}{Peak Signal to Noise Ratio}
\newacronym{mse}{MSE}{Mean Square Error}
\newacronym{lstm}{LSTM}{Long Short-Term Memory}
\newacronym{a2c}{A2C}{Advantage Actor Critic}
\newacronym{rmsd}{RMSD}{Root Mean Squared Deviation}
\newacronym{rnn}{RNN}{Recurrent Neural Network}
\newacronym{jscc}{JSCC}{Joint Source Channel Coding}